\documentclass[letterpaper]{article}
\usepackage{aaai2027}
\usepackage[hyphens]{url}
\usepackage{graphicx}
\usepackage{natbib}
\usepackage{caption}
\usepackage{amsmath}
\usepackage{amssymb}
\usepackage{booktabs}
\usepackage{colortbl}
\usepackage{tabularx}
\usepackage{mathtools}

\usepackage{algorithm}
\usepackage{algorithmic}
\usepackage{cuted}
\usepackage{marvosym}

\definecolor{ourmethodrow}{RGB}{235,243,250}
\title{Energy-Guided Flow Matching}
\author{Haoyang Tong\textsuperscript{* \rm 1,2}, Yu He\textsuperscript{* \rm 2}, Fang Li\textsuperscript{\rm 2}, Lichen Ma\textsuperscript{\rm 2,3}, Jingling Fu\textsuperscript{\rm 2}, Dong Chen\textsuperscript{\rm  2}, Zhen Chen\textsuperscript{\rm 2}\\Junshi Huang\textsuperscript{\rm 2}, Jie Cao\textsuperscript{\Letter \rm 1}}
\affiliations{
    \textsuperscript{\rm 1}MAIS \& NLPR, CASIA \quad\textsuperscript{\rm 2}JD.com \quad\textsuperscript{\rm 3}Xi'an Jiaotong University\\
    \tt\small tonghaoyang22@mails.ucas.ac.cn, heyu2579@gmail.com, jie.cao@cripac.ia.ac.cn

}

\begin{document}
\maketitle
\begingroup
\renewcommand{\thefootnote}{}
\footnotetext{* Equal contribution \qquad
\Letter Corresponding author}
\addtocounter{footnote}{-1}
\endgroup

\begin{abstract}
Pixel-space generative models bypass lossy latent compression, yet necessitate joint learning of global structure and fine-grained details in a high-dimensional space. 
Standard flow matching interpolates noise toward a fixed clean-image endpoint, leaving the spectral evolution to be learned implicitly. 
In this paper, we introduce \emph{Energy-Guided Flow Matching}(EG-FM) that explicitly models a coarse-to-fine generative trajectory by moving endpoint. 
Specifically, EG-FM replaces the fixed endpoint with a heat-kernel-filtered endpoint that evolves smoothly from low-frequency image to clean image.
The fraction of high-frequency signal in moving endpoint is released by an image-specific energy-guided scheduling, leading to the re-targeting of velocity in flow matching.
Our framework requires no adaptation of the backbone and training data, bringing negligible cost on the training and inference stages. 
In our experiment, EG-FM consistently achieves lower FID on the ImageNet class-conditional image generation task at \(256 \times 256\)  with fewer epochs, reaching an FID of 1.55 at 200 epochs and 1.45 at 600 epochs. 
We continue training the generation task on the setting of \(512 \times 512\) resolution, yielding a FID of 1.58 after only 40 high-resolution adaptation epochs.
Furthermore, we transfer EG-FM on text-to-image generation and achieve 0.85 on GenEval score and 83.9 on DPG-Bench. 
Code is available at \textcolor{blue}{\url{https://github.com/ysng123/EG-FM}}.
\end{abstract}

\section{Introduction}

Pixel-space generative models learn image distributions directly on pixels, avoiding information loss caused by latent compression,
and thus preserve fine-grained details of images~\cite{baade2026latentforcing,cai2026hidream}.
Recent advances in diffusion models and flow matching have enabled
high-fidelity image synthesis, reconstruction, and editing
~\cite{ho2020ddpm,lipman2023flow}.
Despite that, high-quality pixel-space generation remains a challenging task due to the high-dimensional space of images and complex dependencies across spatial information, which require the models to coordinate global structure with local high-frequency details.

\begin{figure}[t]
    \centering
    \includegraphics[width=\columnwidth]{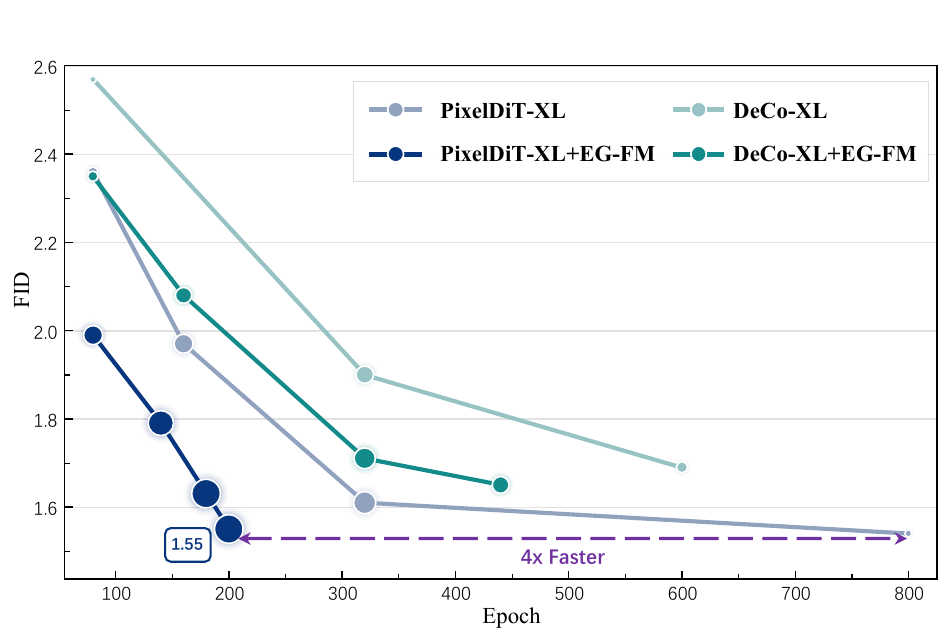}
    \caption{\textbf{Comparison of baselines and Energy-Guided Flow
    Matching (EG-FM).} 
    Marker size increases as checkpoints approach the desirable lower-left region, representing fewer training epochs and lower FID. 
    Based on backbones of PixelDiT and DeCo, the EG-FM variants achieve improved FID performance with significant fewer training epochs.}
    \label{fig:introduction}
\end{figure}

A central problem of pixel-level image generation is how the generative process recovers the components of visual information.  
Existing flow-matching methods typically connect a noise distribution to a fixed clean-image endpoint, assuming that the model should move towards the same target image throughout generation progress~\cite{liu2023rectified,ma2024sit}.
Although mathematically simple and efficient, this formulation does not explicitly model the coarse-to-fine progress of pixel-level image generation.
Generally, low-frequency components responsible for overall structure are usually established at early stage, while high-frequency textures and details emerge later~\cite{rissanen2023heat,hoogeboom2022blurring}. 
Unified modeling on full frequencies may increase the difficulty of task learning~\cite{ma2026deco,lin2026frepix}.
However, this problem may be less severe in latent diffusion models, as the high-frequency components in images are discarded in VAE compression~\cite{rombach2022ldm}.  

Recent pixel-level generation approaches attempt to address this problem through improved network architectures~\cite{chen2025pixelflow,yu2025pixeldit}, stronger patchify strategy~\cite{starodubcev2026registers}, and additional training objectives ~\cite{singh2025irepa,ma2026deco}.
In those works, the study of generative trajectory design is still under-explored.
A path that explicitly integrates the frequency evolution of images could provide an appropriate inductive bias for pixel-level generation, allowing the model to establish the global structure before synthesizing high-frequency details.
However, designing a dynamic trajectory is not straightforward, because its intermediate endpoints must progressively integrate appropriate components of visual information while keeping compatible with the underlying framework of flow-matching.

Inspired by this idea, we revise the pixel-level generation from the perspective of generative-trajectory design and propose \emph{Energy-Guided Flow Matching}.  
This method introduces a moving endpoint to replace the fixed clean endpoint in standard flow matching, allowing the generative process to evolve according to the frequency compositions of each image. 
Specifically, we construct a smooth heat-time scheduling along time-steps and control the information-release rate according to the scheduled energy distribution of each image. 
In this way, EG-FM first recovers low-frequency structure and then generates high-frequency texture and details in a coarse-to-fine manner.
In our experiments, EG-FM consistently improves the FID of image generation tasks on variant backbones with fewer training epochs, as shown in Fig.~\ref{fig:introduction}.

Our main contributions are summarized as follows:
\begin{itemize}
    \item We propose Energy-Guided Flow Matching, a new dynamic generative trajectory that replaces the fixed clean endpoint with a sample-adaptive moving endpoint by progressively releasing frequency information according to the spectral energy of each image.
    
    \item We derive the velocity target induced by the moving
    spectral endpoint and develop a unified energy-based parameterization of the releasing schedule, making EG-FM compatible with standard flow-matching.

    \item Extensive experiments show that EG-FM improves generation quality and training efficiency on various existing backbones.
    It achieves FIDs of 1.45 and 1.58 on ImageNet class-conditional image generation at the resolutions of $256 \times 256$ and $512 \times 512$, respectively, and obtains scores of 0.85 on GenEval and 83.9 on DPG-Bench on text-to-image generation.
\end{itemize}

\section{Related Work}
\label{sec:related-work}

\subsection{Pixel-Space Generative Modeling}

High-resolution pixel-space generation is difficult because dense RGB representations increase optimization complexity and computational cost. ~\cite{li2025jit}
Existing methods mainly address this through two complementary strategies: improving high-resolution optimization through resolution-aware noise schedules~\cite{hoogeboom2023simple}, loss reweighting, and more effective capacity allocation~\cite{hoogeboom2025sid2}; And restructuring cross-scale computation using hourglass attention~\cite{crowson2024hdit}, progressive-resolution generation~\cite{chen2025pixelflow}, or hierarchical global--local architectures~\cite{yu2025pixeldit,chen2026dip,he2026hyperdit}. These advances suggest that scaling pixel-space generative models depends not only on backbone capacity, but also on how computation is distributed across resolutions and how the learning target is parameterized~\cite{li2025jit,guo2026pixelu}.
Despite their effectiveness, these methods typically optimize the model along a predefined probability path, while leaving the temporal emergence of different spatial frequencies to be inferred implicitly from the training data. Our method instead embeds sample-adaptive frequency release directly into the probability path. By explicitly supervising a coarse-to-fine ordering, our formulation reduces the burden on the backbone to discover this ordering solely through optimization. It consequently induces a more structured transition from global layout to fine-grained appearance without increasing model capacity.

\begin{figure*}[t]
    \centering
    \includegraphics[width=0.98\textwidth]{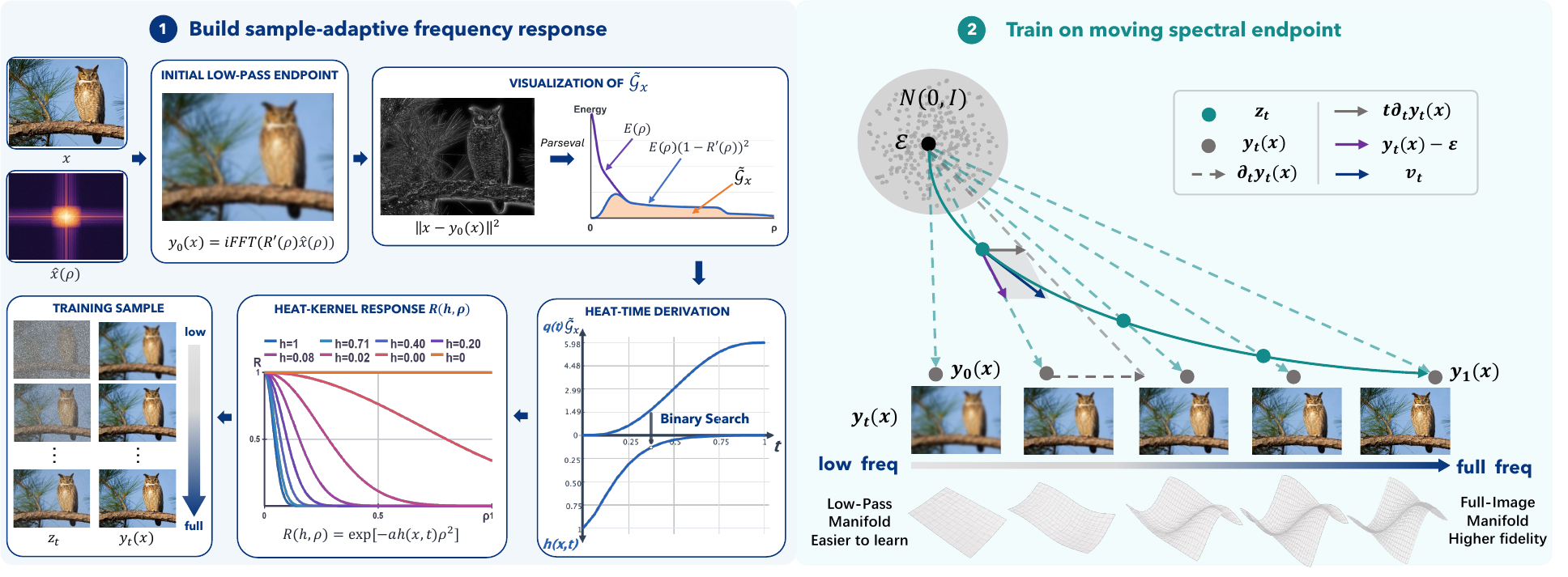}
    \caption{Overview of Energy-Guided Flow Matching.  Starting from a clean
    image, we construct a low-frequency endpoint and measure the residual spectral
    gap $\mathcal{\tilde{G}}_x$. 
    Based on the global release clock $q(t)$, we derive a sample-adaptive heat time $h(x,t)$ that controls the heat-kernel response $R(h,\rho)$, enabling the progressive recovery of frequency components from low to high.
    For the training trajectory, the moving spectral endpoints $y_t(x)$ evolve from a easily predicted low-pass manifold toward the full-image manifold, forming a coarse-to-fine curved path $z_t$.
    The velocity $y_t(x_0)-\epsilon$ toward the current endpoint and the endpoint-motion term $t\,\partial_t y_t(x_0)$ together form the target velocity $v_t$.
    }
    \label{fig:framework}
\end{figure*}

\subsection{Coarse-to-Fine Generation}

Coarse-to-fine generative priors~\cite{lee2022deblurring} are typically implemented by decomposing the data representation or generative process, assigning low-frequency semantics and high-frequency details to different branches~\cite{ma2026deco,ren2026freqflow,ma2026frequencyboosterfullfrequencymodelinghighfidelity}, scales~\cite{zhao2026lapflow}, transport paths~\cite{lin2026frepix}, or auxiliary paths~\cite{he2026liwi}. Through frequency-band decomposition, pyramidal representations, and frequency-aware objectives, these methods improve multiscale modeling and demonstrate that explicit spectral priors can stabilize global structure while facilitating the recovery of local detail.
Recent trajectory-centric methods seek to embed spectral priors into the generative path.\cite{lee2022deblurring,lin2026frepix,zhao2026lapflow} However, they generally use a temporal schedule shared across samples, overlooking cross-image variation in spectral composition. Consequently, the same time may represent different reconstruction progress across samples.
Our method defines a sample-adaptive dynamic path conditioned on each image’s spectral characteristics. Aligning the trajectory with sample-specific spectral composition makes the shared time variable better reflect comparable relative progress, yielding a coarse-to-fine evolution matched to each image’s content. Because this organization is built directly into the probability path, the model can accurately learn generative dynamics from analytically defined intermediate states without iterative trajectory simulation during training.

\section{Method}
\label{sec:method}

In Sec.~\ref{sec:dynamic-endpoint}, we propose the framework of Energy-Guided Flow Matching which integrates the coarse-to-fine prior in the endpoints of generation trajectory.
In Sec.~\ref{sec:adaptive-schedule}, the sample-adaptive frequency components is built through heat-time scheduling to explicitly guide intermediate states evolving from global structure to local details. 
We further derive the learning framework on the target velocity of moving endpoint in Sec.~\ref{sec:dynamic-velocity}.


\subsection{Moving Spectral Endpoint}
\label{sec:dynamic-endpoint}

The noisy sample in standard flow matching is defined as the interpolation between Gaussian noise $\epsilon\sim \mathcal{N}(0,I)$ and a clean endpoint $x\sim p_{\mathrm{data}}$:
\begin{equation}
    z_t=t x+(1-t)\epsilon.
    \label{eq:standard-path}
\end{equation}
Given the fixed image, the direction of velocity points towards the same full-spectrum endpoint $x$ at every time-step $t$.
In this way, the generation processes of global structure and high-frequency details are implicitly modeled within the same trajectory.
To explicitly modulate the frequency components in the supervision, we instead use
\begin{equation}
    z_t=t\,y_t(x)+(1-t)\epsilon,
    \label{eq:dynamic-path}
\end{equation}
where $y_t(x)$ is designed to evolve from a low-pass variant of $x$ to full-spectrum $x$ itself. 
As illustrated in the right part of Fig. ~\ref{fig:framework}, the high-frequency single of $x$ is progressively released to produce gradually sharpened endpoints $y_t(x)$ along time-step $t$.
This design produces a set of moving endpoints that evolves from a coarse structure to fine-grained details.
When $t=1$, we require $y_1(x) = x$.
Since $z_0=\epsilon$ and $z_1=x$, this modification preserves the constraints of boundary distributions in the underlying formulation of flow matching.


To be an effective target from coarse-to-fine, $y_t$ should evolve continuously and release frequency components in a consistent order.
We achieve these properties using a heat-kernel response to generate a smooth, nested family of low-pass images. 
Let $\rho\in[0,1]$ represents the normalized radial frequency, and $\sigma_0$ is the hyper-parameter as discussed later, the heat-kernel frequency response is
\begin{align}
    R'(\rho)&=\exp(-(\pi\sigma_0)^2 * \rho^2).
    \label{eq:initial-response}
\end{align}
Denote  $\widehat{x}(\rho)=\mathcal{F}(x)(\rho)$ the Fourier coefficient of image $x$ at radial frequency $\rho$, we define the initial low-pass image
\begin{align}
    y_0(x)&=\mathcal{F}^{-1}\!\left(
      R'(\rho) * \widehat{x}\right).
    \label{eq:dynamic-initial-endpoint}
\end{align}
where $\mathcal{F}^{-1}$ is the inverse Fourier transform, and the degradation strength of low-pass image $y_0(x)$ can be controlled by the hyper-parameter $\sigma_0$.
To design a set of images with gradually increased high-frequency components, we introduce an image-specific monotonic heat-time $h(x,t) \in [0,1]$, and let  $a=(\pi\sigma_0)^2$ for simplicity. 
The heat-kernel frequency response over the discrete spectrum can be re-formulated as
\begin{align}
    R(h(x,t),\rho)&=\exp(-a * h(x,t) * \rho^2).
    \label{eq:dynamic-response}
\end{align}
By tuning the value of $h(x,t)$, we can obtain heat-kernel responses with different degradation strengths.
Therefore, the set of low-pass images, \textit{i.e.}, moving spectral endpoints, can be re-written as
\begin{align}
    y_t(x)&=\mathcal{F}^{-1}\!\left(
      R(h(x,t),\rho) * \widehat{x}\right).
    \label{eq:dynamic-endpoint}
\end{align}

As the pivotal factor for moving endpoints, we discuss the design principle of heat-time $h(x,t)$.
Given the image $x$, we define $h(x,0)=1$ when $t=0$ to obtain the initial low-pass image $y_0(x)$ since $R(h(x,0),\rho) = R'(\rho)$.
As $t$ increases from 0 to 1, the frequency components are required to be released progressively from low-frequency to high-frequency, until $h(x,1)=0$ reaches the full-spectrum endpoint $y_0(x) = x$, where the information of all frequencies in $x$ is fully released.
Thus, the heat-time $h(x,\cdot)$ is a monotonically decreasing curve as $t$ increases from 0 to 1.


The real demonstration of heat-kernel response curves is presented in the left part of Fig.~\ref{fig:framework}.
The heat-kernel acts as a low-pass filter when $t=0$, and
releases more high-frequency signals as $t$ approaches 1.
Therefore, the set of moving endpoints can be easily obtained via these heat-kernel responses.  
Note that heat-time $h(x,t)$ in heat-kernel is image dependent, we discuss the specific solution to heat-time in Sec.~\ref{sec:adaptive-schedule} for image adaptation. 

\subsection{Energy-Guided Heat-Time Scheduling}
\label{sec:adaptive-schedule}

\begin{figure}[!t]
    \centering
    \includegraphics[width=\columnwidth]{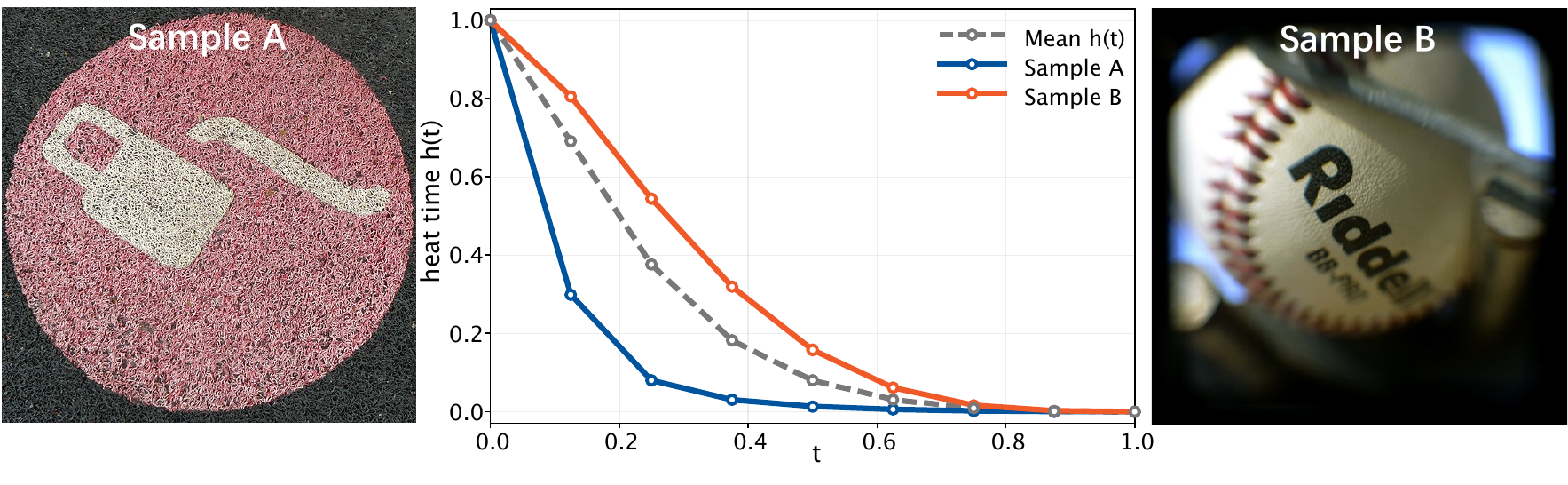}
    \caption{Visualization of sample-adaptive heat-time. Under the same release clock, texture-rich Sample A requires a faster decrease in heat time than Sample B. Lower heat time indicates weaker attenuation and greater frequency release; the dashed curve denotes mean heat-time over ImageNet.}
    \label{fig:adaptive-heat-time}

\end{figure}

Usually, images with rich texture contain much more high-frequency components than that of smooth images, a static heat-time curve is insufficient to manipulate the dynamics in variant images.
We propose to derive $h(x,t)$ based on the deficiency of high-frequency information in endpoint $y_t(x)$ by comparing the information gap between $y_t(x)$ and clean image $x$. 
 

Generally, we define the information gap $\mathcal{\tilde{G}}_x$ as the $\ell_2$ distance from the endpoint $y_0(x)$ to the clean image $x$. 
According to Parseval's theorem, the $\ell_2$ distance of two images can be re-written as this spectral formulation under the setting of unitary Fourier transform:
\begin{equation}
\begin{aligned}
\mathcal{\tilde{G}}_x
    :&= \left\|x-y_0(x)\right\|_2^2 \\
     &\overset{\mathclap{\text{Parseval}}}{=}
     \quad \sum_{\rho}
     \left[ 
        \widehat{x}(\rho)
        -R(h(x,0),\rho)*\widehat{x}(\rho)
     \right]^2 \\
     &= \sum_{\rho}E(\rho)\left[1-R(1,\rho)\right]^2 .
\end{aligned}
\label{eq:total-release}
\end{equation}
where $h(x,0)=1$ and $E(\rho)=\|\widehat{x}(\rho)\|_2^2$ denotes the spectral energy of $x$ at frequency $\rho$.
Intuitively, the information gap $\mathcal{\tilde{G}}_x$ in the frequency-domain consists of the frequency-specific spectral energy  $E(\rho)$ with the corresponding weight $[1-R(1,\rho)]^2$.
Thus, $\mathcal{\tilde{G}}_x$  can be interpreted as the total missing energy for the recovery of high-frequency signal from endpoint $y_0(x)$ to clean image $x$.
Analogously, the information gap between the intermediate endpoint $y_t(x)$ and the initial endpoint $y_0(x)$ can be written as
\begin{equation}
    \mathcal{G}_x(h(x,t))
    =\sum_{\rho}E(\rho)
      \left[R(h(x,t),\rho)-R(1,\rho)\right]^2.
    \label{eq:released-energy}
\end{equation}
At $t=0$ and $h(x,t)=1$, the information gap $\mathcal{G}_x(1)=0$ indicates that the high-frequency energy is the same to low-pass endpoint $y_0(x)$.
As $t$ increases, the heat-kernel frequency response $R(h(x,t),\rho)$ gradually releases the high-frequency components of $x$.
Therefore, $\mathcal{G}_x(h(x,t))$ progressively approaches $\mathcal{\tilde{G}}_x$.
When $t=1$, we get $h(x,t)=0$ and $R(0,\rho)=1$, all the missing high-frequency components has been restored, yielding $y_1(x)=x$ and $\mathcal{G}_x(0)=\mathcal{\tilde{G}}_x$. 
Under the unitary discrete Fourier transform, $\mathcal{G}_x(h)$ is exactly the squared $\ell_2$ distance between the current endpoint $y_t(x)$ and the initial low-pass endpoint $y_0(x)$ and thus directly measures the amount of change along the endpoint moving path.

To ensure that the same time $t$ corresponds to comparable frequency release progress across images, we introduce a global release clock $q(t)$ that aligns the ratio of each image's recovered spectral energy $\mathcal{G}_x(h)$ to its total missing energy $\mathcal{\tilde{G}}_x$:
\begin{equation}
    \mathcal{G}_x(h(x,t))/\mathcal{\tilde{G}}_x=q(t).
    \label{eq:energy-matching}
\end{equation}
As $t$ increases from $0$ to $1$, the recovered spectral energy $\mathcal{G}_x(h(x,t))$ increases from $0$ to the total missing spectral energy $\tilde{\mathcal{G}}_x$. 
Accordingly, $q(t):[0,1]\rightarrow[0,1]$ is a non-decreasing release clock satisfying $q(0)=0$ and $q(1)=1$.
We use the smootherstep function as the default release clock. 
More discussion can be found in experiment.
Consequently, samples at the same $t$ complete the same fraction of energy recovery while retaining sample-adaptive heat-time $h(x,t)$ which is derived from Eq.~\eqref{eq:energy-matching}. 
The resulting $h(x,t)$ adapts the frequency-release rate to each image's spectrum while remaining synchronized by the same release clock. 
Fig.~\ref{fig:adaptive-heat-time} confirms that Eq.\eqref{eq:energy-matching} produces distinct sample-adaptive heat-time under the global release clock. 
Both heat-time eventually approaches zero, and the dashed gray curve denotes the mean heat-time. 
Lower $h(x,t)$ corresponds to weaker attenuation and greater frequency release.

\begin{figure*}[t]
  \centering
  \includegraphics[width=\textwidth]{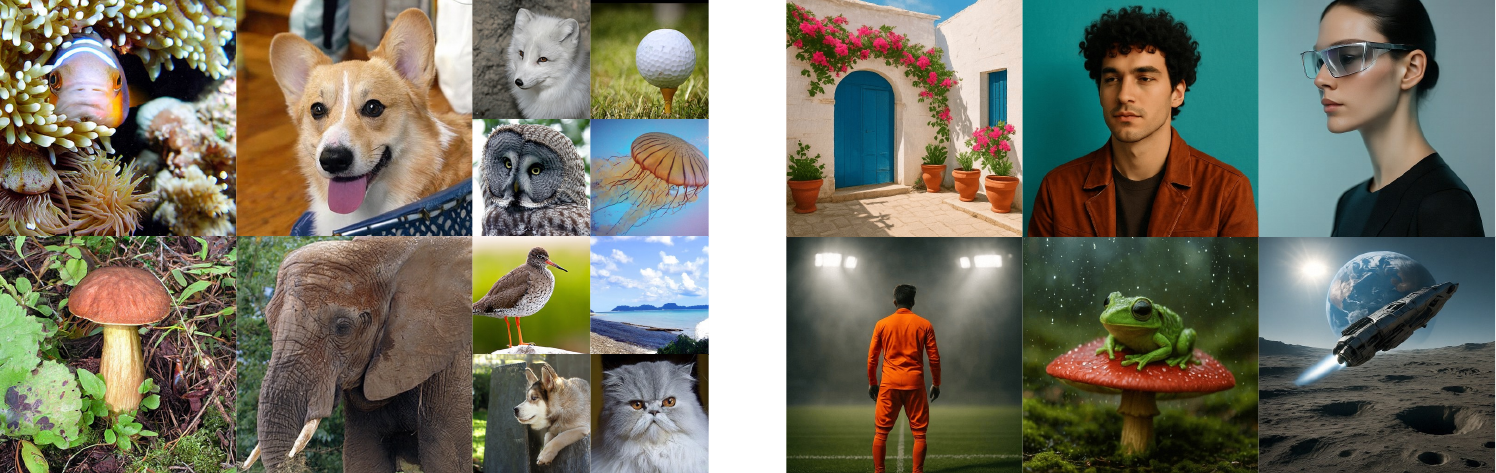}
  \caption{Qualitative results produced by PixelDiT-XL with EG-FM. From left to right, the panels show ImageNet class-conditional samples at $512\!\times\!512$ and $256\!\times\!256$, followed by text-to-image samples at $512\!\times\!512$.}
  \label{fig:qualitative-results}
\end{figure*}

Eq.~\eqref{eq:energy-matching} specifies the desired release clock $q(t)$ rather than the heat-time itself. 
Thus, at each $t$, we recover the sample-specific $h(x,t)$ by inverting $\mathcal{G}_x$. 
A general closed-form inverse is unavailable because $\mathcal{G}_x$ is a weighted sum of squared exponential-response differences; nevertheless, it is straightforward to compute numerically. 
For $h\in[0,1]$, we have $R(h,\rho)\geq R(1,\rho)$, and the response and release derivatives satisfy
\begingroup
\footnotesize
\begin{equation}
    \partial_hR(h,\rho)
    =-a\rho^2R(h,\rho)\leq0,
    \label{eq:R-monotonic-main}
\end{equation}
\begin{equation}
    \partial_h\mathcal{G}_x(h)
    =2\sum_{\rho}E(\rho)
      \left[R(h,\rho)-R(1,\rho)\right]
      \partial_hR(h,\rho)\leq0.
    \label{eq:G-monotonic-main}
\end{equation}
\endgroup
Therefore, $\mathcal{G}_x(h)$ is nonincreasing in $h$ and, for a nondegenerate spectrum, strictly decreasing in the interior. 
Moreover, $\mathcal{G}_x(0)=\mathcal{\tilde{G}}_x$ and $\mathcal{G}_x(1)=0$, while the target $q(t)\mathcal{\tilde{G}}_x$ always lies between these endpoint values. 
The interval $[0,1]$ is therefore a valid bisection bracket for every $t$, and strict monotonicity guaranties a unique feasible root.

\subsection{Energy-Guided Velocity}
\label{sec:dynamic-velocity}

Because the endpoint evolves over time, the target velocity must equal the exact derivative of the path that generates its noisy sample $z_t$, rather than the velocity field used in fixed-endpoint flow matching. 
Differentiating Eq.~\eqref{eq:dynamic-path} yields
\begin{align}
    v_t
    &=\frac{d z_t}{d t}
      =\frac{d}{d t}\left[t\,y_t(x)+(1-t)\epsilon\right] \notag\\
    &=y_t(x)-\epsilon+t\,\partial_t y_t(x).
    \label{eq:dynamic-velocity}
\end{align}
The first term, $y_t(x)-\epsilon$, represents the basic transport velocity from noise $\epsilon$ to endpoint $y_t$.
The second term, $t\,\partial_t y_t(x)$, arises from the endpoint's own motion and contributes to the generated trajectory in proportion to the current interpolation weight $t$. 
The coefficient $t$ controls how strongly endpoint motion influences the generated trajectory: near $t=0$, the endpoint carries little weight, so its motion has limited influence on the state; as $t$ increases, this influence progressively strengthens. 
Together, the two terms form the energy-guided velocity $v_t$. 
The right panel of Fig.~\ref{fig:framework} illustrates this decomposition in the trajectory diagram: the basic transport velocity and the endpoint-motion contribution sum to the true path tangent.

The endpoint motion $t\,\partial_t y_t(x)$ follows by differentiating Eq.~\eqref{eq:dynamic-endpoint} with respect to $t$. 
Applying the chain rule to the heat-kernel response gives
\begin{equation}
    \partial_t y_t(x)
    =\mathcal{F}^{-1}\!\left(
      -a\rho^2R(h(x,t),\rho)\partial_t h(x,t)\widehat{x}
      \right).
    \label{eq:y-dot-main}
\end{equation}
It remains to determine $\partial_t h(x,t)$. 
To do so without differentiating through the iterative solver, we implicitly differentiate the recovered energy ratio in Eq. \eqref{eq:energy-matching}:
\begin{equation}
    \partial_h\mathcal{G}_x(h(x,t))\,\partial_t h(x,t)/\mathcal{\tilde{G}}_x
    =\partial_t q(t)\,.
    \label{eq:implicit-energy-derivative}
\end{equation}

Solving for $\partial_t h(x,t)$ yields
\begin{equation}
    \partial_t h(x,t)
    =\frac{\partial_t q(t)\,\mathcal{\tilde{G}}_x}{\partial_h\mathcal{G}_x(h(x,t))}.
    \label{eq:h-dot-main}
\end{equation}
Substituting Eq.~\eqref{eq:h-dot-main} into Eq.~\eqref{eq:y-dot-main} provides the required endpoint-motion term. 
Detailed derivations are provided in the Appendix A.
We train the velocity predictor $v_\theta$ with Eq.~\eqref{eq:dynamic-path} providing the noisy sample and Eq.~\eqref{eq:dynamic-velocity} providing the target:
\begin{equation}
\resizebox{0.875\linewidth}{!}{$\displaystyle
\mathcal{L}_{\mathrm{FM}}\!=\!\mathbb{E}_{x,\epsilon,t}
\left[
\left\lVert
v_\theta(z_t,t)\!
-\!
\left(
y_t(x)\!-\!\epsilon\!+\!t\partial_t y_t(x)
\right)
\right\rVert_2^2
\right] .
$}
\label{eq:egfm-loss}
\end{equation}
Thus, EG-FM changes only the path and target, keeping the standard flow-matching denoising progress unchanged.


\section{Experiments}
\label{sec:experiments}

We evaluate Energy-Guided FM on class-conditional generation at
$256\!\times\!256$ and $512\!\times\!512$, and on text-to-image generation at
$512\!\times\!512$.  Fig.~\ref{fig:qualitative-results} visualizes image
samples across these settings, providing a qualitative overview.

\begin{table*}[!t]
  \centering
  \small
  \setlength{\tabcolsep}{3.2pt}
  \begin{tabularx}{\textwidth}{@{}l*{8}{>{\centering\arraybackslash}X}@{}}
    \toprule
    Method & Epochs & \#Params & NFE &
      FID $\downarrow$ & sFID $\downarrow$ & IS $\uparrow$ &
      Precision $\uparrow$ & Recall $\uparrow$ \\
    \midrule
    \textcolor{gray}{REPA~\cite{yu2025repa}}         & \textcolor{gray}{800} & \textcolor{gray}{675M} & \textcolor{gray}{250$\times$2} &
      \textcolor{gray}{1.42} & \textcolor{gray}{4.70} & \textcolor{gray}{305.7} &
      \textcolor{gray}{0.80} & \textcolor{gray}{0.65} \\
    \textcolor{gray}{DDT-XL~\cite{wang2025ddt}}       & \textcolor{gray}{400} & \textcolor{gray}{675M} & \textcolor{gray}{--} &
      \textcolor{gray}{1.26} & \textcolor{gray}{--} & \textcolor{gray}{310.6} &
      \textcolor{gray}{0.79} & \textcolor{gray}{0.65} \\
    \textcolor{gray}{RAE-XL~\cite{zheng2025rae}}       & \textcolor{gray}{800} & \textcolor{gray}{839M} & \textcolor{gray}{--} &
      \textcolor{gray}{1.13} & \textcolor{gray}{--} & \textcolor{gray}{262.6} &
      \textcolor{gray}{0.78} & \textcolor{gray}{0.67} \\
    \arrayrulecolor{gray}\midrule\arrayrulecolor{black}
    PixelFlow-XL~\cite{chen2025pixelflow} & 320  & 677M & 120$\times$2 & 1.98 & 5.83 & 282.1 & 0.81 & 0.60 \\
    PixNerd-XL~\cite{wang2025pixnerd}     & 320  & 700M & 100$\times$2 & 1.93 & --   & 298.0 & 0.80 & 0.60 \\
    JiT-G~\cite{li2025jit}                & 600  & 2.0B & 100$\times$2 & 1.82 & --   & 292.6 & 0.79 & 0.62 \\
    PixelU-H/16~\cite{guo2026pixelu}      & 600  & 1.17B& 100$\times$2 & 1.63 & 5.04 & 305.9 & 0.79 & 0.64 \\
    DiP-XL/16~\cite{chen2026dip}          & 600  & 631M & 100$\times$2 & 1.79 & 4.59 & 281.9 & 0.80 & 0.63 \\
    FREPix-XL~\cite{lin2026frepix}         & 320  & 674M & 100$\times$2 & 1.91 & 4.59 & 295.6 & 0.79 & 0.62 \\
    DeCo-XL/16~\cite{ma2026deco}          & 600  & 682M & 100$\times$2 & 1.69 & 4.59 & 304.0 & 0.79 & 0.63 \\
    \rowcolor{ourmethodrow}
    \quad + Energy-Guided FM & 440 & 682M & 100$\times$2 & 1.63 & 4.78 & 300.1 & 0.79 & 0.62 \\
    HyperDiT-H~\cite{he2026hyperdit} & 600  & 952M & 100$\times$2 & 1.56 & 4.73 & 306.5 & 0.80 & 0.64 \\
    \rowcolor{ourmethodrow}
    \quad + Energy-Guided FM & 220 & 952M & 100$\times$2 & 1.51 & 4.31 & 293.4 & 0.78 & 0.64 \\
    PixelDiT-XL~\cite{yu2025pixeldit} & 80   & 797M & 100$\times$2 & 2.36 & 5.11 & 282.3 & 0.80 & 0.57 \\
    PixelDiT-XL~\cite{yu2025pixeldit} & 320  & 797M & 100$\times$2 & 1.61 & 4.68 & 292.7 & 0.78 & 0.64 \\
    PixelDiT-XL~\cite{yu2025pixeldit} & 800 & 797M & 100$\times$2 & 1.54 & 4.49 & 297.0 & 0.78 & 0.65 \\
    \rowcolor{ourmethodrow}
    \quad + Energy-Guided FM & 80  & 797M & 100$\times$2 & 1.99 & 5.09 & 280.8 & 0.81 & 0.61 \\
    \rowcolor{ourmethodrow}
    \quad + Energy-Guided FM & 200 & 797M & 100$\times$2 & 1.55 & 4.60 &
      296.2 & 0.79 & 0.65 \\
    \rowcolor{ourmethodrow}
    \quad + Energy-Guided FM & 600 & 797M & 100$\times$2 & 1.45 & 4.41 & 299.6 & 0.78 & 0.65 \\
    \bottomrule
  \end{tabularx}
  \caption{Class-conditional generation on ImageNet
  $256\!\times\!256$.  Each Energy-Guided FM row uses the same backbone and ADM evaluation protocol.  PixelDiT additionally
  expose convergence at multiple budgets.  NFE denotes the number of function
  evaluations, with $\times 2$ accounting for conditional and unconditional
  CFG evaluations.}
  \label{tab:imagenet-256}
\end{table*}

\begin{table*}[!t]
  \centering
  \begin{minipage}[t]{0.48\textwidth}
    \centering
    \small
    \setlength{\tabcolsep}{2pt}
    \begin{tabularx}{\linewidth}{@{}l*{4}{>{\centering\arraybackslash}X}@{}}
      \toprule
      Method & Epochs & Params & FID $\downarrow$ & IS $\uparrow$ \\
      \midrule
            \textcolor{gray}{DiT-XL/2} &
        \textcolor{gray}{600} & \textcolor{gray}{675M} &
        \textcolor{gray}{3.04} & \textcolor{gray}{240.8} \\
      \textcolor{gray}{SiT-XL/2} &
        \textcolor{gray}{600} & \textcolor{gray}{675M} &
        \textcolor{gray}{2.62} & \textcolor{gray}{252.2} \\
        \textcolor{gray}{REPA} &
        \textcolor{gray}{200} & \textcolor{gray}{675M} &
        \textcolor{gray}{2.08} & \textcolor{gray}{274.6} \\
      \arrayrulecolor{gray}\midrule\arrayrulecolor{black} 
      PixNerd-XL$^{\dagger}$         & 320  & 700M & 2.84 & 245.6 \\
      JiT-H              & 600   & 956M   & 1.94 & 309.1 \\
    PixelU-H/32 & 600 & 1.2B & 1.92 & 322.1 \\
    DiP-XL/32        & -  & 631M & 2.31 & 291.7\\ 
    DeCo-XL/16$^{\dagger}$         & 340  & 682M & 2.22 & 290.0\\
      PixelDiT-XL$^{\dagger}$        & 850  & 797M & 1.81 & 278.6 \\
      \rowcolor{ourmethodrow}
      \quad + EG-FM$^{\dagger}$ & 240 & 797M & 1.68 & 295.5 \\
      \rowcolor{ourmethodrow}
      HyperDiT-H + EG-FM$^{\dagger}$ & 260 & 952M & 1.58 & 285.0 \\
      \bottomrule
    \end{tabularx}
    \captionof{table}{Quantitative comparison for class-conditional generation on ImageNet
    $512\!\times\!512$. ${\dagger}$ denotes continued training from a checkpoint on ImageNet $256\!\times\!256$.}
    \label{tab:imagenet-512}
  \end{minipage}\hfill
  \begin{minipage}[t]{0.48\textwidth}
    \centering
    \small
    \setlength{\tabcolsep}{2pt}
    \begin{tabularx}{\linewidth}{@{}l*{4}{>{\centering\arraybackslash}X}@{}}
      \toprule
      Method & Params & GenEval $\uparrow$ &
        DPG $\uparrow$ \\
      \midrule
      PixArt-$\alpha$                & 0.6B & 0.48 & 71.6 \\
      PixArt-$\Sigma$               & 0.6B & 0.52 & 79.5 \\
      SD3                         & 8B & 0.68 & - \\
      FLUX.1-dev                       & 12B & 0.67 & 82.5 \\
      BLIP3o                          & 4B & 0.81 & 79.4 \\
      OmniGen2                      & 4B & 0.80 & 83.6 \\
      \midrule
      PixelFlow                      & 0.9B & 0.60 & 77.9 \\
      PixNerd                            & 1.2B & 0.73 & 80.9 \\
      DeCo-XXL/16                    & 1.1B & \textbf{0.86} & 81.4 \\
      PixelDiT-T2I                 & 1.3B & 0.78 & 83.7 \\
      \rowcolor{ourmethodrow}
      EG-FM-T2I                 & 1.3B & 0.85 & \textbf{83.9} \\
      \bottomrule
    \end{tabularx}
    \captionof{table}{Quantitative comparison for text-to-image generation at $512 \times 512$ on GenEval and DPG-Bench.}
    \label{tab:t2i-results}
  \end{minipage}
\end{table*}

\subsection{Experimental Setup}
\label{sec:experimental-setup}

\paragraph{Class-conditional generation.}
We train from scratch on ImageNet-1K~\cite{deng2009imagenet} using the official DeCo-XL/16 and PixelDiT-XL
implementations, and reimplement HyperDiT-H following the paper~\cite{he2026hyperdit}.  
For each backbone, the data processing, training hyperparameters and sample method follow the corresponding original paper.  All reported metrics for our ImageNet models are computed from 50K generated samples using the ADM evaluation suite
~\cite{dhariwal2021adm}.  Specifically, we report FID, sFID, Inception Score
(IS), precision, and recall.

\paragraph{Text-to-image.}
Following PixelDiT~\cite{yu2025pixeldit}, we adopt Gemma-2~\cite{team2024gemma} as the text encoder and PixelDiT as the image generation backbone. 
The model is trained on the BLIP3o~\cite{chen2025blip3} dataset using a three-stage training strategy. 
In the first stage, we train the model at $256 \times 256$ with a batch size of 1,024 for 200K steps. 
In the second stage, the resolution is increased to $512 \times 512$, and the model is trained with a batch size of 384 for an additional 100K steps. 
Finally, we perform supervised fine-tuning on the BLIP3o-60K at $512 \times 512$ with a batch size of 384 for 40K steps.
We evaluate on GenEval~\cite{ghosh2023geneval} with 533 prompts and DPG-Bench~\cite{hu2024ella} with 1065 prompts.

\paragraph{Sampling and efficiency protocol.}
For every backbone, we use the sampling method, time discretization, precision,
and number of function evaluations specified by its original implementation.
The Standard-FM and EG-FM runs thus differ neither in solver nor in
sampling budget.  In particular, DeCo and HyperDiT use their original 50-step
Heun samplers, whereas PixelDiT uses its original 100-step FlowDPMSolver.
Unless it is the variable under study, Energy-Guided FM uses
$\sigma_0=3.5$, the quintic smootherstep release clock
$q(t)=6t^5-15t^4+10t^3$, and 16 bisection iterations to solve the
sample-dependent heat time $h(x,t)$.

\subsection{Class-Conditional Image Generation}
\label{sec:c2i-results}

\paragraph{ImageNet $256\!\times\!256$.}
Table~\ref{tab:imagenet-256} shows that Energy-Guided FM consistently improves distributional fidelity across DeCo, HyperDiT, and PixelDiT, while providing a clear convergence advantage. DeCo-XL/16 reaches an FID of 1.63 after 440 epochs, compared with 1.69 after 600 epochs for the baseline. HyperDiT-H achieves an FID of 1.51 after 220 epochs, outperforming the baseline result of 1.56 after 600 epochs. On PixelDiT-XL, EG-FM reduces FID from 2.36 to 1.99 at 80 epochs and reaches 1.55 at 200 epochs, already surpassing the 320-epoch baseline result of 1.61. Continued training further improves FID to 1.45 at 600 epochs. Because the backbone architecture, sampler, and evaluation protocol are held fixed, these controlled comparisons isolate the probability trajectory as the primary experimental difference, demonstrating both faster convergence and a better final FID. The consistent gains across three architecturally distinct backbones further suggest that sample-adaptive coarse-to-fine energy release provides a transferable path prior rather than a backbone-specific optimization heuristic.

\paragraph{ImageNet $512\!\times\!512$.}
The $512\times 512$ experiment evaluates whether the same trajectory remains effective when the number of pixels is quadrupled.  We initialize HyperDiT and PixelDiT from their 220-epoch and 200-epoch checkpoints and fine-tune for only 40 epochs each with a learning rate of $1\times 10^{-5}$.  As shown in Table~\ref{tab:imagenet-512}, PixelDiT with EG-FM reaches an FID of 1.68 and an Inception Score of 295.5, compared with 1.81 and 278.6 for the standard PixelDiT trained for 530 additional epochs. HyperDiT with EG-FM achieves an FID of 1.58, obtaining the best result among the compared methods. These results demonstrate that EG-FM can transfer effectively across resolutions, enabling strong high-resolution performance with limited additional training. The consistent results across two distinct backbones further suggest that energy-guided release provides a scalable path prior rather than a resolution-specific optimization strategy.

\subsection{Text-to-Image Generation}
\label{sec:t2i-results}

Text-to-image evaluation tests whether EG-FM transfers beyond class conditioning. 
This is a stronger transfer test than changing resolution alone because the model must preserve a text-conditioned semantic while releasing pixel-space detail.  
Table~\ref{tab:t2i-results} compares EG-FM-T2I with PixelDiT-T2I~\cite{yu2025pixeldit} and DeCo-XXL/16~\cite{ma2026deco}, and recent latent-space systems evaluated at $512 \times 512$ ~\cite{li2026ppflow}.  
The two metrics capture complementary aspects of text alignment: GenEval~\cite{ghosh2023geneval} emphasizes object-centric composition, whereas DPG-Bench~\cite{hu2024ella} stresses dense-prompt compliance.  
Compared with PixelDiT-T2I~\cite{yu2025pixeldit}, EG-FM-T2I increases GenEval from 0.78 to 0.85 while also improving DPG by 0.2.
It achieves the best DPG score among all compared methods and the second-best GenEval score, trailing DeCo-XXL/16~\cite{ma2026deco} by only 0.01.
Fig. ~\ref{fig:qualitative-results} right shows the visualization results.

\subsection{Ablation and Analysis}
\label{sec:ablations}

All ablations use PixelDiT-XL trained for 80 epochs, with the
optimizer and evaluation setting held fixed.

\begin{figure}[h]
  \centering
  \begin{minipage}[c]{0.50\columnwidth}
    \centering
    \includegraphics[width=\linewidth]{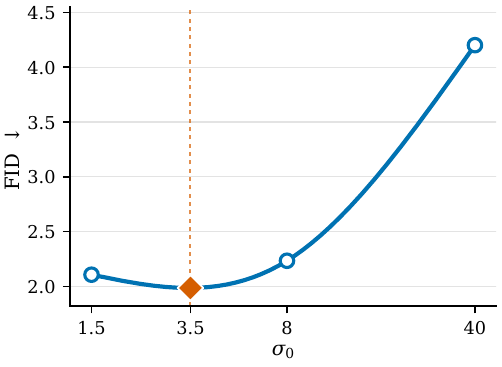}
  \end{minipage}\hfill
  \begin{minipage}[c]{0.48\columnwidth}
    \centering
    \small
    \setlength{\tabcolsep}{2.5pt}
    \begin{tabularx}{\linewidth}{@{}>{\raggedright\arraybackslash}Xcc@{}}
      \toprule
      $\sigma_0$ & FID $\downarrow$ & IS $\uparrow$\\
      \midrule
      0 (Baseline) & 2.36  &282.3\\
      1.5          & 2.11  &268.6\\
      \rowcolor{ourmethodrow}
      3.5 & 1.99 &280.8\\
      8           & 2.23 & 300.4\\
      40          & 4.20 &353.0\\
      1000 (DC only) & 66.08 &40.1\\
      \bottomrule
    \end{tabularx}
  \end{minipage}
  \caption{Initial-endpoint sensitivity after 80
  epochs.  The left curve highlights the nonzero practical range; the right table
  reports full sweep, $\sigma_0=3.5$ gives the best FID.}
  \label{fig:ablation-endpoint}
\end{figure}

\paragraph{Initial endpoint strength.}
Figure~\ref{fig:ablation-endpoint} illustrates how $\sigma_0$ controls the initial endpoint. When $\sigma_0=0$, the endpoint remains unchanged, recovering the baseline. $\sigma_0=1.5$ yields mild suppression of high-frequency components, whereas $\sigma_0=3.5$ achieves the best balance between removing premature fine-grained details and preserving structural guidance. Stronger filtering at $\sigma_0=8$ or $40$ removes excessive spatial information and consequently degrades performance. At $\sigma_0=1000$, the endpoint approaches the DC-only limit. 
The U-shaped shows that creating an intermediate endpoint suppresses premature details while preserving sufficient spatial structure.
We therefore set $\sigma_0=3.5$ by default.

\noindent
\begin{minipage}[t]{\columnwidth}
  \vspace{2pt}
  \centering
  \begin{minipage}[t]{0.48\linewidth}
    \centering
    \small
    \setlength{\tabcolsep}{3pt}
    \begin{tabularx}{\linewidth}{@{}>{\raggedright\arraybackslash}Xcc@{}}
      \toprule
      Release schedule & FID $\downarrow$ & IS $\uparrow$\\
      \midrule
      Shared linear & 2.48 &264.8\\
      Dataset-level & 2.11 & 275.6\\
      Class-level & 2.03 &274.6\\
      \rowcolor{ourmethodrow}
      Sample-level    & 1.99 & 280.8\\
      \bottomrule
    \end{tabularx}
    \captionof{table}{Ablation on the granularity of heat-time.}
    \label{tab:ablation-release-granularity}
  \end{minipage}\hfill
  \begin{minipage}[t]{0.48\linewidth}
    \centering
    \small
    \setlength{\tabcolsep}{3pt}
    \begin{tabularx}{\linewidth}{@{}>{\raggedright\arraybackslash}Xcc@{}}
      \toprule
      $q(t)$ & FID $\downarrow$ & IS $\uparrow$\\
      \midrule
      Linear               & 2.08 &274.1\\
      Smoothstep           & 2.01 &277.3\\
      \rowcolor{ourmethodrow}
      Smootherstep         & 1.99 &280.8\\
      Sigmoid              & 2.03 &278.3\\
      \bottomrule
    \end{tabularx}
    \captionof{table}{Ablation on the release-clock function.}
    \label{tab:ablation-release-allocation}
  \end{minipage}
\end{minipage}
\par

\paragraph{Contribution of adaptive heat-time schedule.}
Table~\ref{tab:ablation-release-granularity} compares four adaptive schedule strategies under the same initial low-pass endpoint and training protocol. `Shared linear' applies $h(t)=1-t$ to all images. `Dataset-level' averages image-specific heat-time schedules over the training set, while `Class-level'' averages them within each class. `Sample-level' derives heat time $h(x,t)$ from each image's spectral energy. Performance improves from shared linear to dataset-level, class-level, and sample-level variant, indicating that increasingly fine-grained spectral adaptation is beneficial. The best performance of the sample-level variant confirms sample adaptivity as a key component of EG-FM.

\paragraph{Effect of release-clock curvature.}
Table~\ref{tab:ablation-release-allocation} compares linear, smoothstep, smootherstep, and sigmoid function. 
Detailed function formulations are provided in the appendix C.3. 
All four functions share the same boundary conditions and total release but allocate energy differently over time. The comparison keeps the initial endpoint and per-sample energy target unchanged.  The quintic smootherstep
gives the best FID at 80 epochs and is used in all main experiments.  
Since the compared functions share the same endpoints and total released energy,
their ordering shows that generation quality depends not only on what
frequency content is released, but also on when it is introduced along the
path.

\noindent
\begin{minipage}[t]{\columnwidth}
  \vspace{2pt}
  \centering
  \begin{minipage}[t]{0.48\linewidth}
    \centering
    \includegraphics[width=\linewidth]{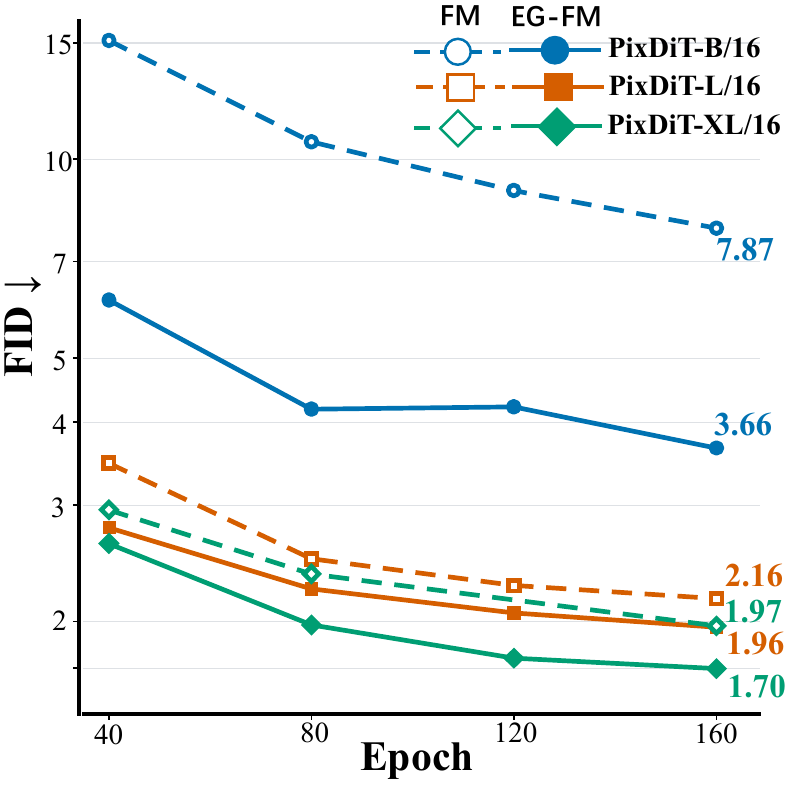}
    \captionof{figure}{Generality of EG-FM across model sizes.}
    \label{fig:ablation-modelsize}
  \end{minipage}\hfill
  \begin{minipage}[t]{0.48\linewidth}
    \centering
    \includegraphics[width=\linewidth]{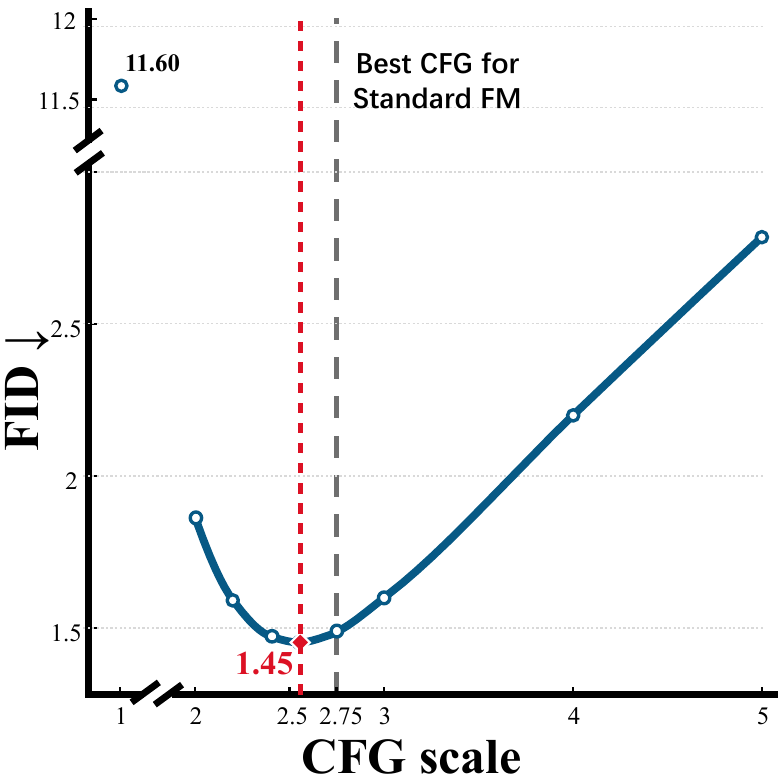}
    \captionof{figure}{Sweep on CFG scale.}
    \label{fig:ablation-cfg-summary}
  \end{minipage}
\end{minipage}
\par







\paragraph{Generality across model sizes.}
We further assess whether the effectiveness of EG-FM depends on model capacity by evaluating it across three model scales. As shown in Fig.~\ref{fig:ablation-modelsize}, EG-FM consistently outperforms the corresponding Standard-FM baseline at every scale. These results demonstrate that EG-FM provides a general trajectory design that transfers across model capacities, rather than an optimization tailored to a particular model size.

\paragraph{CFG scale.}
We tune the CFG scale for EG-FM to account for its modified generative trajectory. As shown in Figure~\ref{fig:ablation-cfg-summary}, the CFG ablation yields a U-shaped FID curve. Without guidance, the FID is $11.60$; it decreases steadily as the guidance scale increases, reaches a minimum of $1.45$ at CFG $=2.55$, and then rises again. We therefore use a CFG scale of $2.55$ for EG-FM, slightly below the original setting.
\section{Conclusion}
\label{sec:conclusion}

We introduced \textbf{Energy-Guided Flow Matching (EG-FM)}, a sample-adaptive trajectory for pixel-space generation that replaces the fixed clean endpoint with a heat-kernel-filtered endpoint. 
By aligning the ratio of each image recovered spectral energy to its total missing energy, a sample-adaptive heat-time is derived to govern the same frequency release rate across images.
Extensive experiments show that EG-FM consistently improves generation quality and training efficiency across class-conditional and text-to-image generation without architectural changes or additional losses. 
These results validate Energy-Guided Flow Matching as a effective design for pixel-space image generation.

\bibliography{aaai2027}

\clearpage
\appendix
\section{Detailed Derivation of EG-FM}
\label{app:derivation}

This section provides the full derivation of the sample-adaptive heat-time
schedule and its path-consistent velocity target.

\subsection{Spectral Endpoint Path}
\label{app:spectral-path}

Let $x\sim p_{\mathrm{data}}$ be a clean image and
$\epsilon\sim\mathcal{N}(0,I)$ be a Gaussian source sample.  Standard flow
matching uses the linear conditional path
\begin{equation}
    z_t=t x+(1-t)\epsilon,
    \qquad
    \frac{d z_t}{d t}=x-\epsilon.
    \label{eq:app-standard-path}
\end{equation}
To expose a coarse-to-fine ordering, we replace the fixed clean endpoint by a
moving spectral endpoint
\begin{equation}
    y_t(x)=\mathcal{F}^{-1}\!\left(
        R(h(x,t),\rho)\widehat{x}
    \right),
    \label{eq:app-moving-endpoint}
\end{equation}
where
\begin{equation}
    R(h,\rho)=\exp(-a h\rho^2),
    \qquad
    a=(\pi\sigma_0)^2.
    \label{eq:app-response}
\end{equation}
Here $\rho\in[0,1]$ is the normalized radial frequency and
$\widehat{x}(\rho)=\mathcal{F}(x)(\rho)$.  We use $h(x,0)=1$ and
$h(x,1)=0$.  Consequently, the initial response is
$R(1,\rho)=\exp(-a\rho^2)$, whereas $R(0,\rho)=1$.  Thus, $y_0(x)$ is a
low-pass image and $y_1(x)=x$.  The resulting conditional path is
\begin{equation}
    z_t=t\,y_t(x)+(1-t)\epsilon.
    \label{eq:app-dynamic-path}
\end{equation}
It retains the original boundary distributions because $z_0=\epsilon$ and
$z_1=x$, while changing the intermediate transport geometry.

\subsection{Energy-Matched Heat Time}
\label{app:heat-time}

Let $E(\rho)=\|\widehat{x}(\rho)\|_2^2$ denote the spectral energy of $x$ at
frequency $\rho$, as in the main text.  Under a unitary discrete Fourier
transform, Parseval's identity gives the total squared displacement from the
initial low-pass endpoint to the clean image:
\begin{align}
    \mathcal{\tilde{G}}_x
    &:=\left\|x-y_0(x)\right\|_2^2 \notag\\
    &=\sum_{\rho}E(\rho)
      \left[1-R(1,\rho)\right]^2.
    \label{eq:app-total-gap}
\end{align}
At an arbitrary heat time $h$, the spectral change already recovered from the
initial endpoint is
\begin{equation}
    \mathcal{G}_x(h)
    :=\sum_{\rho}E(\rho)
      \left[R(h,\rho)-R(1,\rho)\right]^2.
    \label{eq:app-release-function}
\end{equation}
The two boundary values are
\begin{equation}
    \mathcal{G}_x(1)=0,
    \qquad
    \mathcal{G}_x(0)=\mathcal{\tilde{G}}_x.
    \label{eq:app-release-boundaries}
\end{equation}
Let $q:[0,1]\rightarrow[0,1]$ be a differentiable, nondecreasing release
clock satisfying
\begin{equation}
    q(0)=0,
    \qquad
    q(1)=1.
    \label{eq:app-release-clock}
\end{equation}
The sample-dependent heat time is defined implicitly by
\begin{equation}
    \mathcal{G}_x(h(x,t))
    =q(t)\mathcal{\tilde{G}}_x.
    \label{eq:app-energy-constraint}
\end{equation}
Therefore, the same path time corresponds to the same fraction of recovered
spectral change for every sample, even though the absolute heat time remains
sample-dependent.

For completeness, differentiating the response with respect to $h$ gives
\begin{equation}
    \partial_h R(h,\rho)
    =-a\rho^2R(h,\rho).
    \label{eq:app-response-h-derivative}
\end{equation}
Substituting Eq.~\eqref{eq:app-response-h-derivative} into the derivative of
Eq.~\eqref{eq:app-release-function} yields
\begin{align}
    \partial_h\mathcal{G}_x(h)
    &=
    2\sum_{\rho}E(\rho)
    [R(h,\rho)-R(1,\rho)]\partial_hR(h,\rho)
    \notag\\
    &=
    \scalebox{0.93}{$\displaystyle
    -2a\sum_{\rho}E(\rho)\rho^2R(h,\rho)
    [R(h,\rho)-R(1,\rho)]\leq 0.$}
    \label{eq:app-release-h-derivative}
\end{align}
For a nondegenerate image spectrum, the inequality is strict in the interior.
Together with Eq.~\eqref{eq:app-release-boundaries}, this proves that
Eq.~\eqref{eq:app-energy-constraint} has a unique solution in $[0,1]$ for
every $t$.  It can therefore be recovered reliably by bracketed bisection.

\subsection{Implicit Derivative and Exact Velocity}
\label{app:exact-velocity}

Differentiating Eq.~\eqref{eq:app-energy-constraint} with respect to path time
gives
\begin{equation}
    \partial_h\mathcal{G}_x(h(x,t))
    \,\partial_t h(x,t)
    =\partial_t q(t)\,\mathcal{\tilde{G}}_x.
    \label{eq:app-implicit-differentiation}
\end{equation}
Hence,
\begin{equation}
    \partial_t h(x,t)
    =
    \frac{\partial_t q(t)\,\mathcal{\tilde{G}}_x}
    {\partial_h\mathcal{G}_x(h(x,t))}.
    \label{eq:app-h-dot-general}
\end{equation}
Using Eq.~\eqref{eq:app-release-h-derivative}, the derivative can be written
as
\begin{equation}
    \partial_t h(x,t)
    =
    \frac{\partial_t q(t)\,\mathcal{\tilde{G}}_x}
    {-2a\sum_{\rho}E(\rho)\rho^2R(h,\rho)
    [R(h,\rho)-R(1,\rho)]},
    \label{eq:app-h-dot-expanded}
\end{equation}
where $h=h(x,t)$.  The numerator is nonnegative and the denominator is
nonpositive, so $\partial_t h(x,t)\leq0$, as required for progressively
weakening attenuation.

The endpoint derivative follows from the chain rule:
\begin{align}
    \partial_t y_t(x)
    &=
    \mathcal{F}^{-1}\!\left(
      \partial_tR(h(x,t),\rho)\widehat{x}
    \right) \notag\\
    &=
    \mathcal{F}^{-1}\!\left(
      -a\rho^2R(h(x,t),\rho)
      \partial_t h(x,t)\widehat{x}
    \right).
    \label{eq:app-endpoint-dot}
\end{align}
Finally, differentiating the actual state path in
Eq.~\eqref{eq:app-dynamic-path} produces
\begin{align}
    v_t
    =\frac{d z_t}{d t}
    &=y_t(x)-\epsilon+t\,\partial_t y_t(x).
    \label{eq:app-exact-velocity}
\end{align}
We use Eq.~\eqref{eq:app-exact-velocity} as the regression target for
velocity-prediction backbones.  If the selected release clock has zero
derivative at the boundaries, the endpoint motion also vanishes smoothly at
both ends of the path.

\subsection{Endpoint Stability}
\label{app:endpoint-stability}

We distinguish two related issues: boundedness of the continuous-time target
and stable floating-point evaluation of that target.  The latter requires
special care near $t=0$, where both the numerator and denominator of
Eq.~\eqref{eq:app-h-dot-general} vanish.  For the analytic result, assume a
nondegenerate spectrum, $q(t)>0$ for $t\in(0,1)$, and the endpoint-flatness
conditions
\begin{equation}
    \lim_{t\downarrow0}\frac{\partial_t q(t)}{\sqrt{q(t)}}=0,
    \qquad
    \lim_{t\uparrow1}\partial_t q(t)=0.
    \label{eq:app-clock-flatness}
\end{equation}
These conditions are stated directly in terms of the general release clock
and do not require a particular functional form.

\paragraph{Proposition 1 (stable endpoints).}
Under Eq.~\eqref{eq:app-clock-flatness}, $y_t(x)$ and $v_t$ extend
continuously to both endpoints, $\partial_t y_t(x)\to0$ as $t\to0,1$, and
\begin{equation}
    z_0=\epsilon,\quad z_1=x,\qquad
    v_0=y_0(x)-\epsilon,\quad
    v_1=x-\epsilon.
    \label{eq:app-stable-endpoints}
\end{equation}

\emph{Proof.}
Taylor expansion at $h=1$ gives
\begin{equation}
  \begin{aligned}
    \mathcal{G}_x(h)
    &=a^2(1-h)^2
      \sum_{\rho}E(\rho)\rho^4R(1,\rho)^2\\
    &\quad+O((1-h)^3),\\
    \partial_h\mathcal{G}_x(h)
    &=-2a^2(1-h)
      \sum_{\rho}E(\rho)\rho^4R(1,\rho)^2\\
    &\quad+O((1-h)^2).
  \end{aligned}
    \label{eq:app-initial-expansion}
\end{equation}
Combining Eq.~\eqref{eq:app-initial-expansion} with the energy constraint
yields the sharper asymptotic relations
\begin{align}
    1-h
    &=
    \sqrt{\frac{\mathcal{\tilde{G}}_x}
    {a^2\sum_{\rho}E(\rho)\rho^4R(1,\rho)^2}}\sqrt{q(t)}
    +O(q(t)), \notag\\
    \partial_t h
    &=
    -\frac{\partial_t q(t)}{2}
    \sqrt{\frac{\mathcal{\tilde{G}}_x}
    {a^2q(t)\sum_{\rho}E(\rho)\rho^4R(1,\rho)^2}}
    \notag\\
    &\quad\times
    \left[1+O\!\left(\sqrt{q(t)}\right)\right].
    \label{eq:app-initial-h-asymptotic}
\end{align}
Thus the apparent $0/0$ in Eq.~\eqref{eq:app-h-dot-general} has a finite,
indeed vanishing, limit.  The first condition in
Eq.~\eqref{eq:app-clock-flatness} and
Eq.~\eqref{eq:app-endpoint-dot} therefore imply
$\partial_t y_t(x)\to0$.

At the other endpoint, $\partial_h\mathcal{G}_x(0)<0$.  Expanding at $h=0$
gives
\begin{equation}
    \mathcal{G}_x(h)
    =\mathcal{\tilde{G}}_x+\partial_h\mathcal{G}_x(0)h+O(h^2).
    \label{eq:app-terminal-expansion}
\end{equation}
Consequently, as $t\uparrow1$,
\begin{equation}
  \begin{aligned}
    h
    &=-\frac{\mathcal{\tilde{G}}_x}
      {\partial_h\mathcal{G}_x(0)}[1-q(t)]
      +O([1-q(t)]^2),\\
    \partial_t h
    &=\frac{\mathcal{\tilde{G}}_x}
      {\partial_h\mathcal{G}_x(0)}
      \partial_t q(t)[1+O(h)].
  \end{aligned}
    \label{eq:app-terminal-h-asymptotic}
\end{equation}
The second condition in Eq.~\eqref{eq:app-clock-flatness} again gives
$\partial_t y_t(x)\to0$.  Substitution into
Eqs.~\eqref{eq:app-dynamic-path} and \eqref{eq:app-exact-velocity} proves
Eq.~\eqref{eq:app-stable-endpoints}.  In particular, both the state and its
velocity remain bounded in endpoint neighborhoods, so the probability-flow
ODE has no endpoint singularity induced by the energy-matched schedule.

\paragraph{Endpoint stability and numerical evaluation in operation..}
For $q(t)=6t^5-15t^4+10t^3$, the endpoint-flatness conditions can be checked
without evaluating a ratio of small quantities.  In fact,
\begin{equation}
  q(t)=t^3(10-15t+6t^2),\qquad
  \partial_t q(t)=30t^2(1-t)^2,
  \label{eq:app-smootherstep-factorization}
\end{equation}
and hence
\begin{equation}
  \frac{\partial_t q(t)}{\sqrt{q(t)}}
  =
  \frac{30\sqrt{t}(1-t)^2}{\sqrt{10-15t+6t^2}}
  \longrightarrow0
  \quad\text{as }t\downarrow0.
  \label{eq:app-smootherstep-stable-ratio}
\end{equation}
Similarly,
$1-q(t)=(1-t)^3[10-15(1-t)+6(1-t)^2]$.
Thus $\partial_t h=O(\sqrt{t})$ at the initial endpoint and
$\partial_t h=O((1-t)^2)$ at the terminal endpoint.  These explicit rates
also show that the endpoint-motion contribution
$t\,\partial_t y_t(x)$ vanishes at both ends.

In the implementation, the two endpoint regions are treated separately.
For $t\leq10^{-5}$ and $1-t\leq10^{-5}$, we evaluate the corresponding
endpoint expressions directly, with the analytic assignments
$(h,\partial_t h)=(1,0)$ at $t=0$ and $(h,\partial_t h)=(0,0)$ at $t=1$.
Outside these regions, $\partial_t h$ is evaluated using the general
expression.  All scalar spectral reductions, root finding, and derivative
calculations are performed in FP32, including under mixed-precision
training.  This endpoint-aware evaluation avoids floating-point \(0/0\)
and suppresses spurious numerical velocity spikes.

\section{Extension to $x$-Prediction}
\label{app:x-prediction}

\subsection{Path-Consistent Parameterization}
\label{app:x-pred-parameterization}

EG-FM can also be applied when the backbone predicts the clean image rather
than velocity.  Let
\begin{equation}
    \widetilde{x}=x_\theta(z_t,t)
    \label{eq:app-x-prediction}
\end{equation}
be the predicted clean image.  Define the energy-guided endpoint operator
\begin{equation}
    \mathcal{T}_t(x)
    :=
    \mathcal{F}^{-1}\!\left(
      R(h(x,t),\rho)\widehat{x}(\rho)
    \right),
    \label{eq:app-release-operator}
\end{equation}
where $h(x,t)$ is computed from the spectrum of its argument using the same
energy-matching rule as Eq.~\eqref{eq:app-energy-constraint}.  The endpoint
implied by the clean-image prediction is then
\begin{equation}
    \widetilde{y}_t=\mathcal{T}_t(\widetilde{x}).
    \label{eq:app-predicted-endpoint}
\end{equation}
Holding $\widetilde{x}$ fixed while applying the explicit path-time
dependence of $\mathcal{T}_t$, its endpoint motion is
\begin{equation}
    \partial_t\widetilde{y}_t
    =
    \left.\partial_t\mathcal{T}_t(x)\right|_{x=\widetilde{x}}.
    \label{eq:app-predicted-endpoint-dot}
\end{equation}
This derivative is evaluated analytically using
Eqs.~\eqref{eq:app-h-dot-general} and \eqref{eq:app-endpoint-dot}; it does not
require differentiating the neural network with respect to time.

Under the EG-FM path, the noise prediction implied by
$\widetilde{x}$ satisfies
\begin{equation}
    z_t=t\widetilde{y}_t+(1-t)\widetilde{\epsilon},
    \qquad
    \widetilde{\epsilon}
    =\frac{z_t-t\widetilde{y}_t}{1-t}.
    \label{eq:app-implied-noise}
\end{equation}
Substituting this result into the dynamic-path velocity gives
\begin{align}
    \widetilde{v}_\theta(z_t,t)
    &=
    \widetilde{y}_t
    +t\,\partial_t\widetilde{y}_t
    -\widetilde{\epsilon} \notag\\
    &=
    t\,\partial_t\mathcal{T}_t(\widetilde{x})
    +\frac{\mathcal{T}_t(\widetilde{x})-z_t}{1-t}.
    \label{eq:app-x-to-v}
\end{align}
Equation~\eqref{eq:app-x-to-v} is the path-consistent conversion from
$x$-prediction to velocity prediction.  Applying the usual fixed-endpoint
conversion would omit both the spectral transformation and its endpoint-motion
term, and would therefore be inconsistent with the states used during EG-FM
training.  At the exact terminal boundary, the clean prediction is used
directly; in numerical sampling Eq.~\eqref{eq:app-x-to-v} is evaluated only at
nonterminal solver times.

\subsection{JiT Result and Discussion}
\label{app:jit-x-prediction}

\begin{center}
    \centering
    \small
    \setlength{\tabcolsep}{7pt}
    \begin{tabular}{lc}
        \toprule
        JiT $x$-prediction path & FID $\downarrow$ \\
        \midrule
        Standard FM & 2.37 \\
        \rowcolor{ourmethodrow}
        {}+ EG-FM & 2.33 \\
        \bottomrule
    \end{tabular}
    \captionof{table}{Controlled $x$-prediction experiment with
    JiT.  EG-FM improves FID from 2.37 to 2.33 under the same
    backbone and evaluation setting.}
    \label{tab:jit-x-prediction}
\end{center}

Table~\ref{tab:jit-x-prediction} shows that the proposed path also transfers to
JiT's $x$-prediction parameterization, reducing FID from 2.37 to 2.33.  The
gain is positive but smaller than those observed with direct velocity
prediction.  The difference follows from what is available when the heat time
is evaluated..  Predictions at early solver
times remain noisy and spectrally inaccurate; consequently,
$E_{\widetilde{x}}(\rho)$ and the inferred $h(\widetilde{x},t)$ can be
unreliable precisely when global structure is first being established.  This mismatch weakens the intended sample-adaptive frequency
ordering and explains the limited 0.04 FID improvement.  In contrast, a
velocity-prediction model directly learns the path-consistent target in
Eq.~\eqref{eq:app-exact-velocity} and does not need to reconstruct a new
sample-specific heat time from an uncertain clean-image estimate at every
sampling step.

\section{Experimental Details}
\label{app:experimental-details}

\begin{figure*}[!t]
  \centering
  \includegraphics[width=\textwidth,trim=0 10bp 0 0,clip]
    {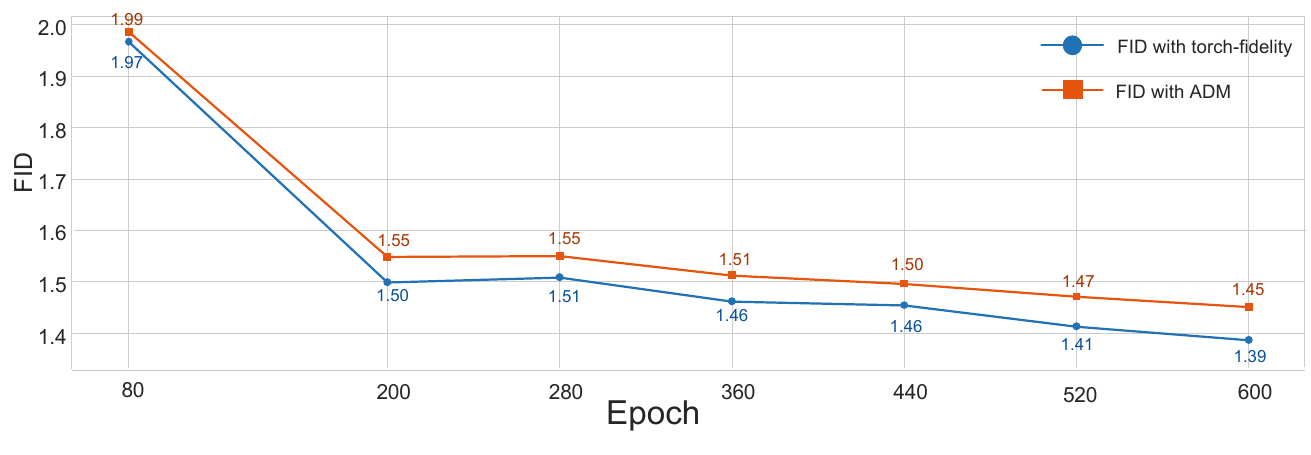}
  \addtocounter{figure}{1}
  \caption{FID across training checkpoints computed with torch-fidelity and
  the ADM evaluation suite.  The two implementations show closely aligned
  convergence trends from 80 to 600 epochs.}
  \label{fig:app-fid-evaluator-trend}
  \addtocounter{figure}{-2}
  \medskip

  \begin{minipage}[t]{0.48\textwidth}
  \hrule height.8pt depth0pt
  \kern2pt
  \captionsetup{position=top,justification=raggedright,
    singlelinecheck=false,labelfont=bf,labelsep=colon,skip=2pt}
  \captionof{algorithm}{Sample-adaptive heat-time solver}
  \label{alg:heat-time-solver}
  \hrule
  \kern2pt
  \begin{algorithmic}[1]
  \REQUIRE $\widehat{x}$, $t\in(0,1)$, $a$, $q$, $K$
  \ENSURE $(h_t,\dot h_t)$
  \STATE $E(\rho)\leftarrow\|\widehat{x}(\rho)\|_2^2$,\quad
  $R_1(\rho)\leftarrow\exp(-a\rho^2)$
  \STATE $\widetilde{\mathcal{G}}_x\leftarrow
  \sum_\rho E(\rho)[1-R_1(\rho)]^2$
  \STATE $(\ell,u)\leftarrow(0,1)$,\quad
  $s\leftarrow q(t)\widetilde{\mathcal{G}}_x$
  \FOR{$k=1,\ldots,K$}
      \STATE $m\leftarrow(\ell+u)/2$,\quad
      $R_m(\rho)\leftarrow\exp(-am\rho^2)$
      \STATE $\mathcal{G}_x(m)\leftarrow
      \sum_\rho E(\rho)[R_m(\rho)-R_1(\rho)]^2$
      \IF{$\mathcal{G}_x(m)>s$}
          \STATE $\ell\leftarrow m$
      \ELSE
          \STATE $u\leftarrow m$
      \ENDIF
  \ENDFOR
  \STATE $h_t\leftarrow(\ell+u)/2$,\quad
  $R_t(\rho)\leftarrow\exp(-a h_t\rho^2)$
  \STATE $D_t\leftarrow-2a\sum_\rho E(\rho)\rho^2R_t(\rho)
  [R_t(\rho)-R_1(\rho)]$
  \STATE $\dot h_t\leftarrow
  q'(t)\widetilde{\mathcal{G}}_x/D_t$
  \end{algorithmic}
  \kern2pt
  \hrule
  \end{minipage}
  \hfill
  \begin{minipage}[t]{0.48\textwidth}
  \hrule height.8pt depth0pt
  \kern2pt
  \captionsetup{position=top,justification=raggedright,
    singlelinecheck=false,labelfont=bf,labelsep=colon,skip=2pt}
  \captionof{algorithm}{EG-FM training procedure}
  \label{alg:egfm-training}
  \hrule
  \kern2pt
  \begin{algorithmic}[1]
  \REQUIRE $\mathcal{B}=\{(x_i,c_i)\}_{i=1}^{B}$, $v_\theta$, $\sigma_0$, $q$, $K$
  \STATE $a\leftarrow(\pi\sigma_0)^2$,\quad $\mathcal{L}\leftarrow0$
  \FOR{$(x,c)\in\mathcal{B}$}
      \STATE $t\sim\mathcal{U}(0,1)$,\quad
      $\epsilon\sim\mathcal{N}(0,I)$,\quad
      $\widehat{x}\leftarrow\mathcal{F}(x)$
      \STATE $(h_t,\dot h_t)\leftarrow
      \mathsf{HeatTime}_q(\widehat{x},t;a,K)$
      \STATE $R_t(\rho)\leftarrow\exp(-a h_t\rho^2)$
      \STATE $y_t\leftarrow\mathcal{F}^{-1}(R_t\widehat{x})$
      \STATE $\dot y_t\leftarrow\mathcal{F}^{-1}
      (-a\rho^2R_t\dot h_t\,\widehat{x})$
      \STATE $z_t\leftarrow t y_t+(1-t)\epsilon$
      \STATE $v_t^\star\leftarrow y_t-\epsilon+t\dot y_t$
      \STATE $\mathcal{L}\leftarrow\mathcal{L}
      +\|v_\theta(z_t,t,c)-v^\star_t\|_2^2/B$
  \ENDFOR
  \STATE $\theta\leftarrow\operatorname{Update}
  (\theta,\nabla_\theta\mathcal{L})$
  \end{algorithmic}
  \kern2pt
  \hrule
  \end{minipage}
\end{figure*}

\subsection{Detailed Results on ImageNet $512\!\times\!512$}
\label{app:detailed-imagenet-512}

Table~\ref{tab:app-imagenet-512-detailed} expands the high-resolution
comparison with sFID, precision, recall, and
sampling cost.  Together with FID and Inception Score, these metrics measure
spatial fidelity, sample quality, and distributional coverage.

\begin{table*}[p]
  \centering
  \small
  \renewcommand{\arraystretch}{0.92}
  \setlength{\abovecaptionskip}{3pt}
  \setlength{\belowcaptionskip}{0pt}
  \setlength{\tabcolsep}{3.2pt}
  \begin{tabularx}{\textwidth}{@{}l*{8}{>{\centering\arraybackslash}X}@{}}
    \toprule
    Method & Epochs & \#Params & NFE & FID $\downarrow$ &
      sFID $\downarrow$ & IS $\uparrow$ & Precision $\uparrow$ &
      Recall $\uparrow$ \\
    \midrule
    \textcolor{gray}{DiT-XL/2} & \textcolor{gray}{600} &
      \textcolor{gray}{675M} & \textcolor{gray}{$250{\times}2$} &
      \textcolor{gray}{3.04} & \textcolor{gray}{5.02} &
      \textcolor{gray}{240.8} & \textcolor{gray}{\textbf{0.84}} &
      \textcolor{gray}{0.54} \\
    \textcolor{gray}{SiT-XL/2} & \textcolor{gray}{600} &
      \textcolor{gray}{675M} & \textcolor{gray}{$250{\times}2$} &
      \textcolor{gray}{2.62} & \textcolor{gray}{\textbf{4.18}} &
      \textcolor{gray}{252.2} & \textcolor{gray}{\textbf{0.84}} &
      \textcolor{gray}{0.57} \\
    \textcolor{gray}{REPA} & \textcolor{gray}{200} &
      \textcolor{gray}{675M} & \textcolor{gray}{$250{\times}2$} &
      \textcolor{gray}{2.08} & \textcolor{gray}{\underline{4.19}} &
      \textcolor{gray}{274.6} & \textcolor{gray}{0.83} &
      \textcolor{gray}{0.58} \\
    \arrayrulecolor{gray}\midrule\arrayrulecolor{black}
    PixNerd-XL$^\dagger$ & 320 & 700M &
      $100{\times}2$ & 2.84 & 5.95 & 245.6 & 0.80 & 0.59 \\
    JiT-H & 600 & 956M &
      $100{\times}2$ & 1.94 & -- & \underline{309.1} & -- & -- \\
    PixelU-H/32 & 600 & 1.2B &
      $100{\times}2$ & 1.92 & 5.98 & \textbf{322.1} & 0.80 & 0.58 \\
    DiP-XL/32 & -- & 631M &
      $100{\times}2$ & 2.31 & 4.48 & 291.7 & \textbf{0.84} & 0.58 \\
    DeCo-XL/16$^\dagger$ & 340 & 682M &
      $100{\times}2$ & 2.22 & 4.67 & 290.0 & 0.80 & 0.60 \\
    PixelDiT-XL$^\dagger$ & 850 & 797M &
      $100{\times}2$ & 1.81 & 5.61 & 278.6 & 0.78 & \textbf{0.67} \\
    \rowcolor{ourmethodrow}
    \quad + EG-FM$^\dagger$ & 240 & 797M & $100{\times}2$ &
      \underline{1.68} & 4.77 & 295.5 & 0.79 & 0.63 \\
    \rowcolor{ourmethodrow}
    HyperDiT-H + EG-FM$^\dagger$ & 260 & 952M & $100{\times}2$ &
      \textbf{1.58} & 4.90 & 285.0 & 0.79 & 0.64 \\
    \bottomrule
  \end{tabularx}
  \caption{Detailed comparison for class-conditional ImageNet generation at
  $512\!\times\!512$.  Metrics are computed on 50K generated samples with the
  ADM evaluation suite.  NFE includes conditional and unconditional
  classifier-free-guidance evaluations; $\dagger$ denotes continued training
  from a $256\!\times\!256$ checkpoint.  A dash indicates an unreported item.}
  \label{tab:app-imagenet-512-detailed}
\end{table*}

\subsection{Detailed Text-to-Image Results}
\label{app:detailed-text-to-image}

Tables~\ref{tab:app-geneval-detailed} and
\ref{tab:app-dpg-detailed} decompose the aggregate text-to-image scores.  GenEval separates object rendering, counting,
color, spatial relation, and attribute-binding capabilities.  DPG-Bench
separately evaluates global consistency, entities, attributes, relations, and
other dense-prompt requirements.

\begin{table*}[p]
  \centering
  \small
  \renewcommand{\arraystretch}{0.92}
  \setlength{\abovecaptionskip}{3pt}
  \setlength{\belowcaptionskip}{0pt}
  \setlength{\tabcolsep}{3.5pt}
  \begin{tabularx}{\textwidth}{@{}l*{8}{>{\centering\arraybackslash}X}@{}}
    \toprule
    Method & \#Params & Single obj. & Two obj. & Counting & Colors &
      Position & Color attr. & Overall $\uparrow$ \\
    \midrule
    PixArt-$\alpha$ & 0.6B & 0.98 & 0.50 & 0.44 & 0.80 & 0.08 & 0.07 & 0.48 \\
    SD3 & 8B & 0.98 & 0.84 & 0.66 & 0.74 & 0.40 & 0.43 & 0.68 \\
    FLUX.1-dev & 12B & \underline{0.99} & 0.81 & \textbf{0.79} &
      0.74 & 0.20 & 0.47 & 0.67 \\
    DALL-E 3 & -- & 0.96 & 0.87 & 0.47 & 0.83 & 0.43 & 0.45 & 0.67 \\
    BLIP3o & 4B & -- & -- & -- & -- & -- & -- & 0.81 \\
    OmniGen2 & 4B & \textbf{1.00} & \textbf{0.95} & 0.64 & 0.88 &
      0.55 & 0.76 & 0.80 \\
    PixelFlow & 0.9B & -- & -- & -- & -- & -- & -- & 0.60 \\
    PixNerd & 1.2B & 0.97 & 0.86 & 0.44 &
      0.83 & \underline{0.71} & 0.53 & 0.73 \\
    DeCo-XXL/16 & 1.1B & \textbf{1.00} & 0.92 &
      0.72 & \underline{0.91} & \textbf{0.80} &
      \textbf{0.79} & \textbf{0.86} \\
    PixelDiT-T2I & 1.3B & \textbf{1.00} &
      0.94 & 0.70 & 0.90 & 0.53 & 0.65 & 0.78 \\
    \rowcolor{ourmethodrow}
    EG-FM-T2I & 1.3B & \textbf{1.00} & \textbf{0.95} & \underline{0.74} & \textbf{0.92} & 0.72 & \underline{0.77} & \underline{0.85} \\
    \bottomrule
  \end{tabularx}
  \caption{Category-wise GenEval results for text-to-image generation at
  $512\!\times\!512$.  Overall is the unweighted mean of the six task scores.
  Bold and underlined values denote the best and second-best results in each
  column.  A dash indicates an unreported item.}
  \label{tab:app-geneval-detailed}
\end{table*}

\begin{table*}[p]
  \centering
  \small
  \renewcommand{\arraystretch}{0.92}
  \setlength{\abovecaptionskip}{3pt}
  \setlength{\belowcaptionskip}{0pt}
  \setlength{\tabcolsep}{5pt}
  \begin{tabularx}{\textwidth}{@{}l*{7}{>{\centering\arraybackslash}X}@{}}
    \toprule
    \multicolumn{1}{l}{Method} & \#Params & Global & Entity & Attribute & Relation & Other &
      Overall $\uparrow$ \\
    \midrule
    PixArt-$\alpha$ & 0.6B & 81.7 & 80.1 & 80.4 & 81.7 & 76.5 & 71.6 \\
    PixArt-$\Sigma$ & 0.6B & 87.5 & 87.1 & 86.5 & 84.0 &
      86.1 & 79.5 \\
    PixelFlow & 0.9B & -- & -- & -- & -- & -- & 77.9 \\
    PixNerd & 1.2B & 80.5 & 87.9 &
      87.2 & \textbf{91.3} & 72.8 & 80.9 \\
    DeCo-XXL/16 & 1.1B & -- & -- & -- & -- & -- & 81.4 \\
    PixelDiT-T2I & 1.3B &
      \underline{88.0} & \textbf{90.9} & \underline{87.6} &
      89.8 & \underline{88.5} & \underline{83.7} \\
    \rowcolor{ourmethodrow}
    EG-FM-T2I & 1.3B & \textbf{89.3} & \underline{89.2} & \textbf{90.2} & \underline{90.9} & \textbf{89.7} & \textbf{83.9} \\
    \bottomrule
  \end{tabularx}
  \caption{Category-wise DPG-Bench results for text-to-image generation at
  $512\!\times\!512$.  Bold and underlined values denote the best and
  second-best results in each column.  A dash indicates an unreported item.}
  \label{tab:app-dpg-detailed}
\end{table*}

\begin{table*}[p]
  \centering
  \small
  \renewcommand{\arraystretch}{0.92}
  \setlength{\abovecaptionskip}{3pt}
  \setlength{\belowcaptionskip}{0pt}
  \setlength{\tabcolsep}{2.5pt}
  \begin{tabular*}{\textwidth}{@{\extracolsep{\fill}}llcccccccc@{}}
    \toprule
    Backbone & Path & Res. & Batch
      & GFLOPs
      & $\Delta$GFLOPs
      & FLOPs inc.
      & Time/step
      & Time inc.
      & Epoch time \\
    \midrule
    DeCo-XL/16     & Standard FM & $256^2$ & 256 & 734.81  & 0.00 & 0.00   & 0.10 & 0.00  & 0.14 \\
    DeCo-XL/16     & EG-FM       & $256^2$ & 256 & 734.86  & 0.06 & $+0.01$ & 0.10 & $+0.41$ & 0.14 \\
    PixelDiT-B/16  & Standard FM & $256^2$ & 256 & 226.29  & 0.00 & 0.00   & 0.04 & 0.00  & 0.06 \\
    PixelDiT-B/16  & EG-FM       & $256^2$ & 256 & 226.34  & 0.06 & $+0.03$ & 0.04 & $+4.81$ & 0.06 \\
    PixelDiT-L/16  & Standard FM & $256^2$ & 256 & 683.30  & 0.00 & 0.00   & 0.09 & 0.00  & 0.12 \\
    PixelDiT-L/16  & EG-FM       & $256^2$ & 256 & 683.36  & 0.06 & $+0.01$ & 0.09 & $+0.51$ & 0.12 \\
    PixelDiT-XL/16 & Standard FM & $256^2$ & 256 & 933.58  & 0.00 & 0.00   & 0.11 & 0.00  & 0.15 \\
    PixelDiT-XL/16 & EG-FM       & $256^2$ & 256 & 933.64  & 0.06 & $+0.01$ & 0.11 & $+0.99$ & 0.15 \\
    PixelDiT-XL/16 & Standard FM & $512^2$ & 64  & 4056.72 & 0.00 & 0.00   & 0.12 & 0.00  & 0.66 \\
    PixelDiT-XL/16 & EG-FM       & $512^2$ & 64  & 4056.98 & 0.26 & $+0.01$ & 0.12 & $+0.92$ & 0.66 \\
    \bottomrule
  \end{tabular*}
  \caption{Matched training compute and wall time.
  GFLOPs and $\Delta$GFLOPs are reported per sample, with
  $\Delta$GFLOPs measured relative to Standard FM.  FLOPs increase and time
  increase are percentage changes relative to the matched Standard FM
  configuration; negative time values indicate faster steps.  Per-step wall
  time is in seconds, and epoch time is in hours. All experiments were conducted on 8 $\times$ B200 GPU.}
  \label{tab:app-training-efficiency}
\end{table*}

\subsection{Release-Clock Ablation}
\label{app:release-clock-ablation}

The release-clock ablation compares four monotone clocks:
\begin{equation}
  \begin{aligned}
    q_{\mathrm{linear}}(t) &= t,\\
    q_{\mathrm{smooth}}(t) &= 3t^2-2t^3,\\
    q_{\mathrm{smoother}}(t) &= 6t^5-15t^4+10t^3,\\
    q_{\mathrm{sigmoid}}(t)
    &=
    \frac{\operatorname{sigm}(k(t-\frac{1}{2}))
    -\operatorname{sigm}(-k/2)}
    {\operatorname{sigm}(k/2)-\operatorname{sigm}(-k/2)},
    \quad k=10,
  \end{aligned}
  \label{eq:app-release-clock-variants}
\end{equation}
where $\operatorname{sigm}(u)=(1+e^{-u})^{-1}$.  The normalization makes
every clock satisfy $q(0)=0$ and $q(1)=1$, so the clocks recover the same
total spectral gap and differ only in temporal allocation.

\begin{center}
\begin{minipage}{0.96\columnwidth}
  \centering
  \includegraphics[width=\linewidth]
    {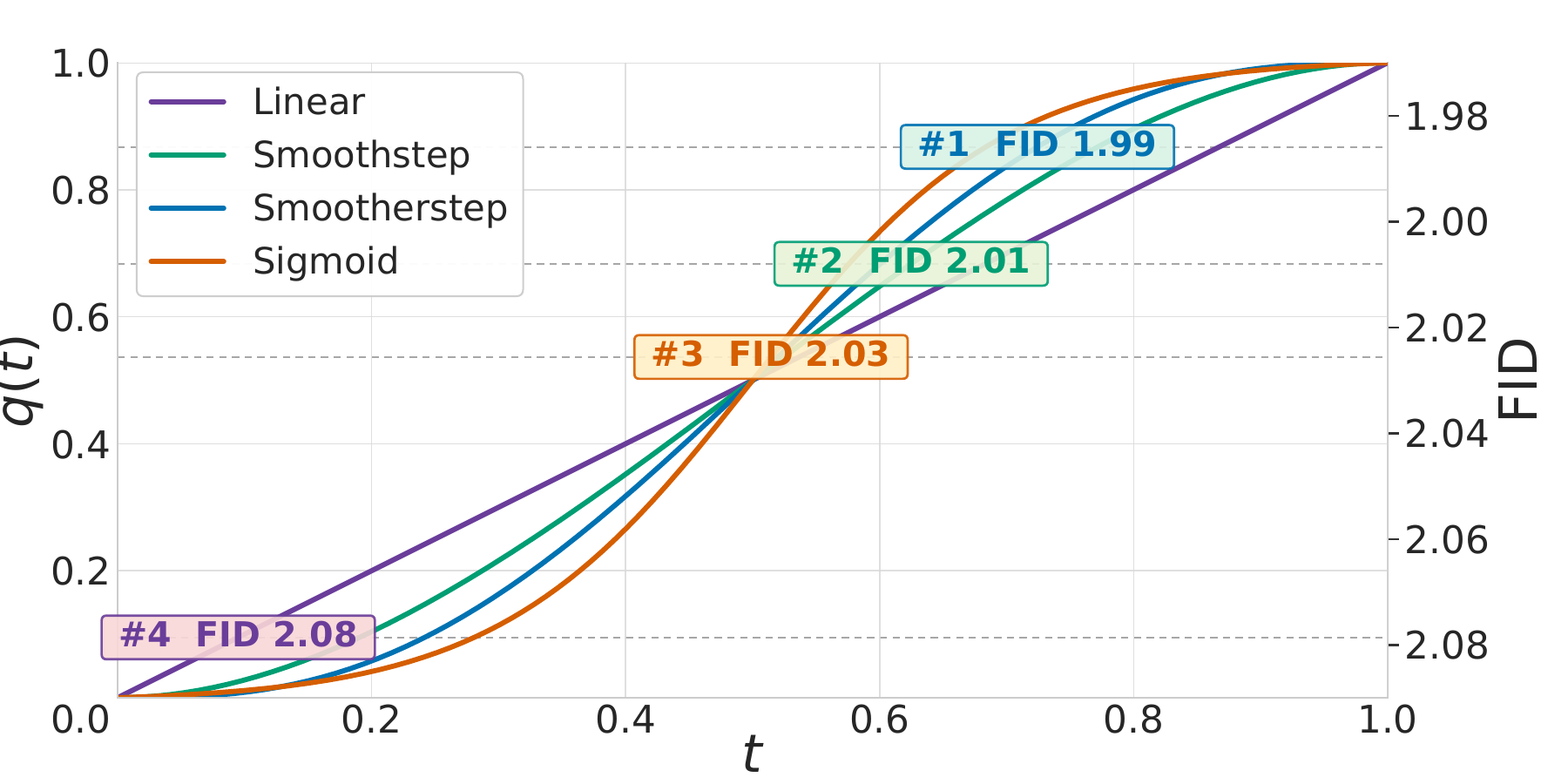}
  \captionof{figure}{Visualization of the release-clock ablation.  Curves show
  the recovered-energy fraction $q(t)$, and annotations report the
  corresponding 80-epoch FID.  Quintic smootherstep achieves
  the best FID (1.99).}
  \label{fig:app-release-clock-ablation}
\end{minipage}
\end{center}

Figure~\ref{fig:app-release-clock-ablation} makes the temporal-allocation
difference explicit.  Linear release allocates progress uniformly, whereas
the three nonlinear clocks delay early recovery and concentrate it near the
middle of the path.  Smootherstep provides the best result while retaining
zero first derivatives at both endpoints, consistent with the stability
analysis above.

\subsection{Training and Inference Efficiency}
\label{app:training-efficiency}
\label{app:inference-efficiency}
We compare Standard FM and EG-FM under identical hardware, global batch size,
numerical precision, data pipeline, gradient-accumulation setting, and logging
frequency. 
Table~\ref{tab:app-training-efficiency} reports the training resolution,
global batch size, per-sample compute, per-step wall-clock time, and per-epoch
wall-clock time for DeCo-XL/16 and three PixelDiT model sizes.

The additional target construction is negligible relative to the backbone
compute: the largest measured increase is only $0.0256\%$ per sample.
Across all matched runs, the per-step wall-clock change ranges from
$+0.41\%$ to $+4.81\%$.  Apart from PixelDiT-B/16, whose measured overhead
Across all matched runs, the per-step wall-clock overhead ranges from
$0.41\%$ to $4.81\%$.  Apart from PixelDiT-B/16, whose measured overhead
is $4.81\%$, all configurations remain within $1\%$ of Standard FM.
Per-epoch times follow the same pattern; consequently, EG-FM's lower epoch
requirement to reach a target FID still translates into lower end-to-end
training time.

At inference, all main DeCo and PixelDiT experiments use direct velocity
prediction.  EG-FM evaluates the learned velocity field without filtering,
FFT or inverse FFT, energy computation, or bisection.  It uses the same
backbone, solver, time grid, and number of function evaluations as Standard
FM.  Consequently, EG-FM does not change inference FLOPs or wall time.

\subsection{EG-FM Algorithms}
\label{app:training-algorithm}

Algorithm~\ref{alg:heat-time-solver} computes the sample-adaptive heat time
$h_t=h(x,t)$ and its derivative $\dot h_t$ using bracketed bisection and
implicit differentiation for $t\in(0,1)$ and
$\widetilde{\mathcal{G}}_x>0$.

\par\medskip
Both algorithms are used only to form training targets.  At inference, direct
velocity prediction integrates $d z_t/dt=v_\theta(z_t,t,c)$ from Gaussian
noise using the protocol described in
Section~\ref{app:training-efficiency}.

\subsection{FID Evaluator Consistency}
\label{app:fid-evaluator-consistency}

Recent works such as JiT report ImageNet FID using torch-fidelity, which can produce slightly lower absolute values than the ADM evaluation suite used for our main results. To facilitate comparison with results reported under either convention, we evaluate the same PixelDiT-XL checkpoints using both implementations. As shown in Figure~\ref{fig:app-fid-evaluator-trend}, the two curves exhibit the same convergence behavior: FID drops sharply between 80 and 200 epochs and then improves steadily through 600 epochs. Torch-fidelity consistently yields values that are only $0.02$--$0.06$ lower, indicating a small evaluator-dependent offset that does not affect the observed training trend. Under the torch-fidelity evaluation protocol, our method achieves an FID of 1.39, further demonstrating state-of-the-art performance in pixel-level generation.

\section{Limitations and Future Work}
\label{app:limitations}

EG-FM has not yet been evaluated on heterogeneous or temporally extended signals, including joint text-image modeling, video generation and embodied decision making, where endpoint guidance must coordinate spatial structure with
temporal dynamics and action.  Its behavior with the largest recent foundation
backbones, including Flux- and Qwen-Image-scale models, also remains untested.
Future work will generalize endpoint-guided flow matching across these
modalities and scales, with the broader goal of developing a unified adaptive
path construction for perception, generation, and control.

\FloatBarrier

\section{Additional Visualizations}
\label{app:additional-visualizations}

The reported checkpoints and evaluation sampling configurations are used for
all additional samples.

\subsection{Class-Conditional Generation}
\label{app:class-conditional-visualizations}

Additional ImageNet samples from PixelDiT-XL with EG-FM at $256^2$ and
$512^2$ cover diverse categories and layouts while preserving global class
structure and detailed textures.  Additional class-conditional sample grids
are shown below.  Within each grid, the two large samples at the top are
generated at $512^2$, while all samples below are generated at $256^2$.

\subsection{Text-to-Image Generation}
\label{app:text-to-image-visualizations}

Additional $512^2$ results cover diverse subjects, attributes, artistic
styles, counting, and spatial relations.  The prompt-aligned samples below
further illustrate EG-FM's compatibility with text conditioning and its
ability to recover fine visual details.

\clearpage
\onecolumn
\begin{figure}[p]
  \centering
  \refstepcounter{figure}
  \label{fig:app-t2i-samples-1}
  \includegraphics[height=\textheight,keepaspectratio]
    {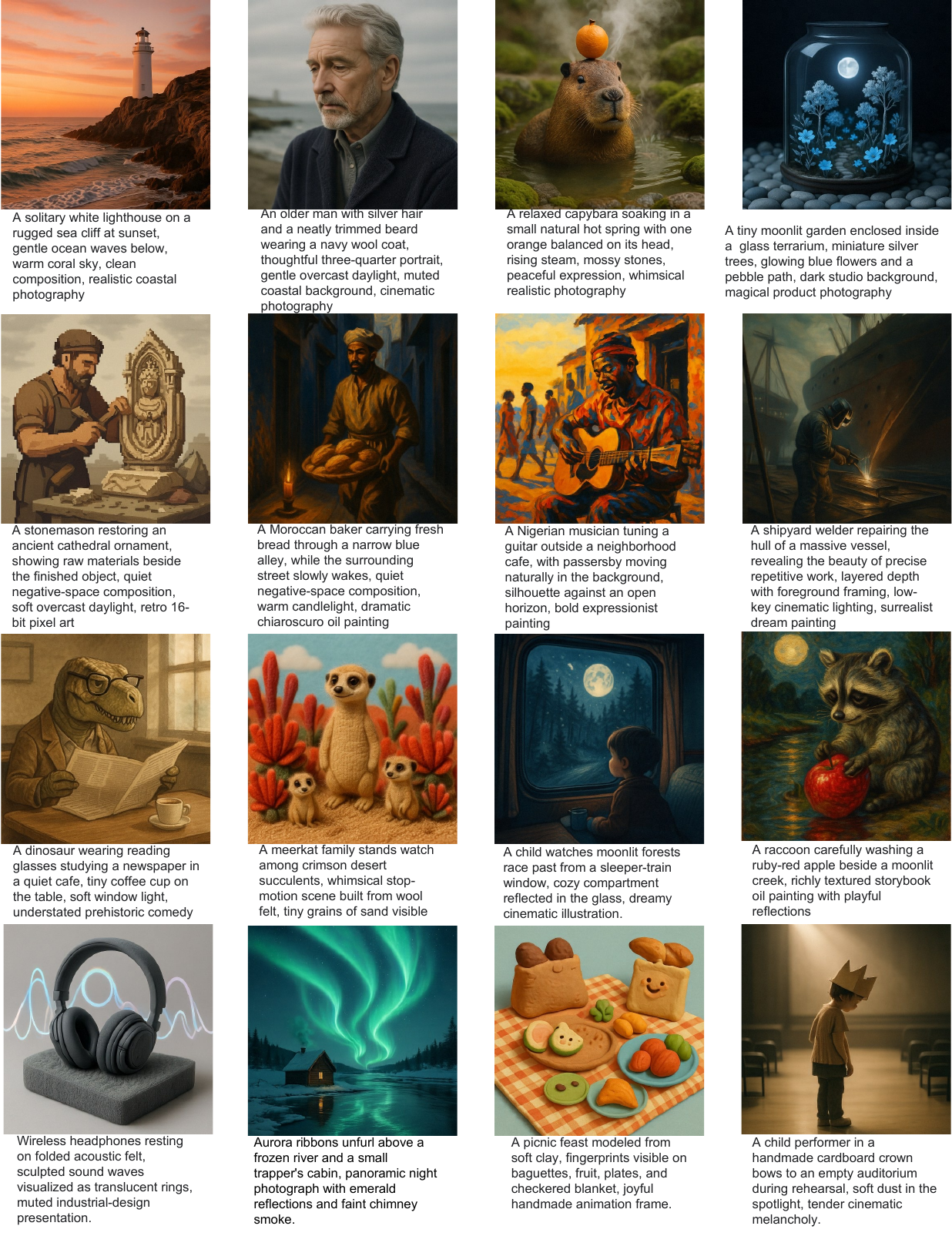}
\end{figure}

\begin{figure}[p]
  \centering
  \captionsetup{skip=0pt}
  \includegraphics[width=0.9641\textwidth,keepaspectratio]
    {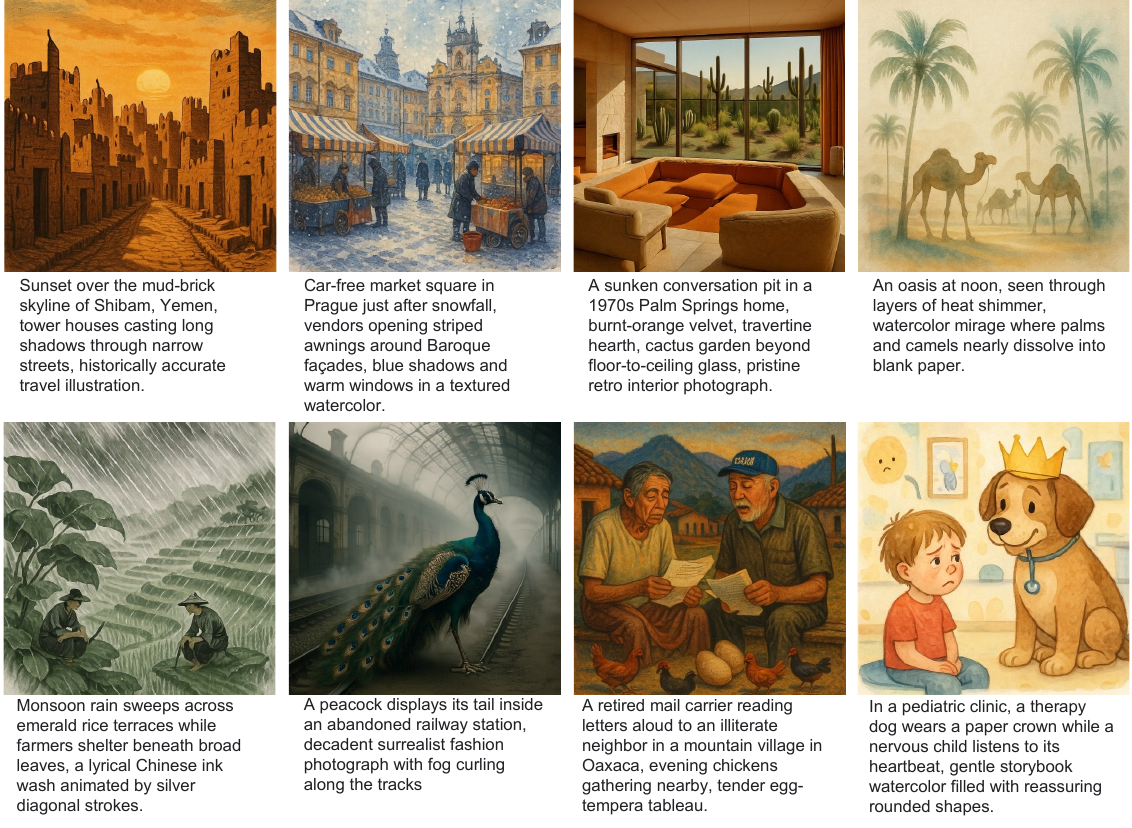}
  \caption{Additional text-to-image samples generated by EG-FM-T2I at
  $512\!\times\!512$, spanning architectural scenes, weather, interiors,
  animals, and human-centered compositions.}
  \label{fig:app-t2i-samples-2}
  \vspace{0pt}
  \includegraphics[width=0.9641\textwidth,keepaspectratio]
    {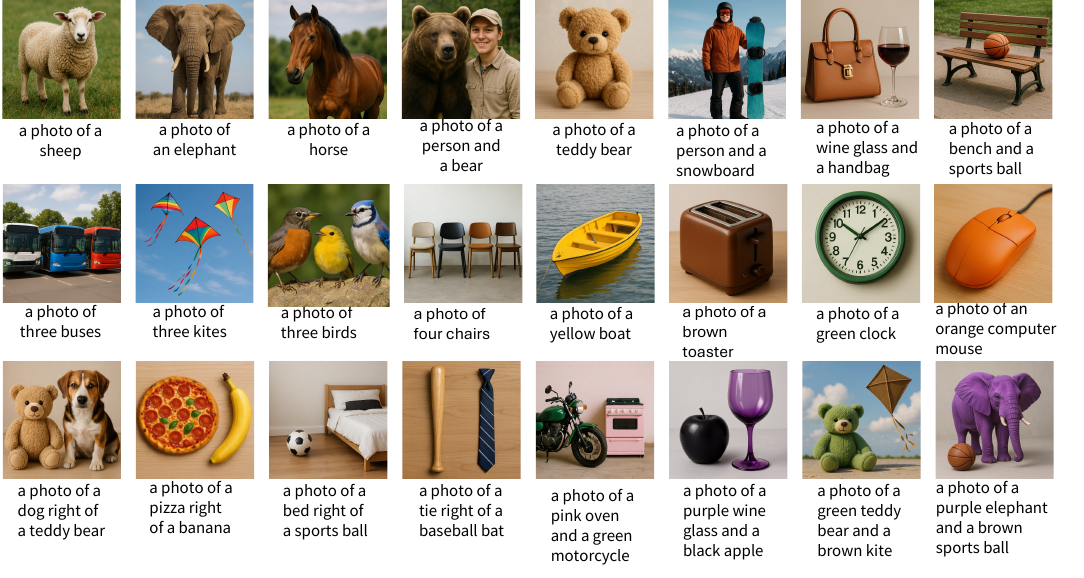}
  \caption{Representative GenEval samples generated by EG-FM-T2I.}
  \label{fig:app-t2i-geneval}
\end{figure}

\clearpage
\vspace*{\fill}
\noindent
  \begin{minipage}[t]{0.48\textwidth}
    \centering
    \includegraphics[width=\linewidth]
      {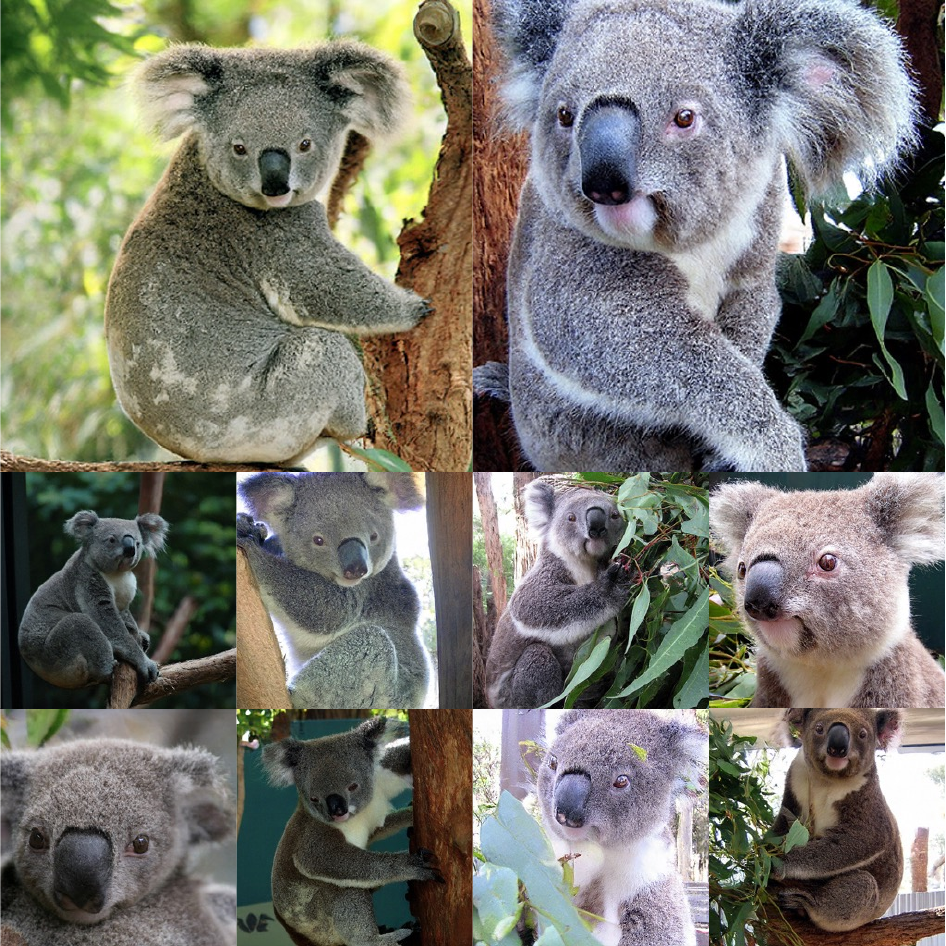}
    \captionof{figure}{ImageNet class 105.}
    \label{fig:app-class-sample-105}
  \end{minipage}\hfill%
  \begin{minipage}[t]{0.48\textwidth}
    \centering
    \includegraphics[width=\linewidth]
      {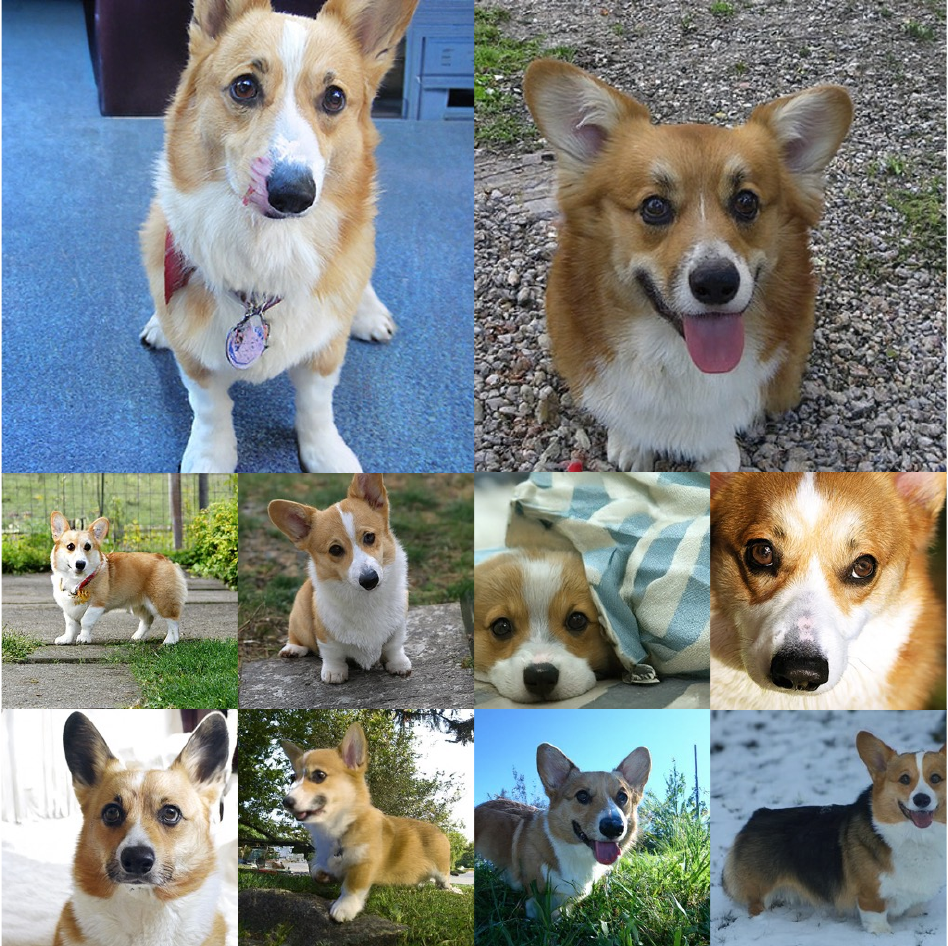}
    \captionof{figure}{ImageNet class 263.}
    \label{fig:app-class-sample-263}
  \end{minipage}

  \par\vspace{0.025\textheight}
\noindent
  \begin{minipage}[t]{0.48\textwidth}
    \centering
    \includegraphics[width=\linewidth]
      {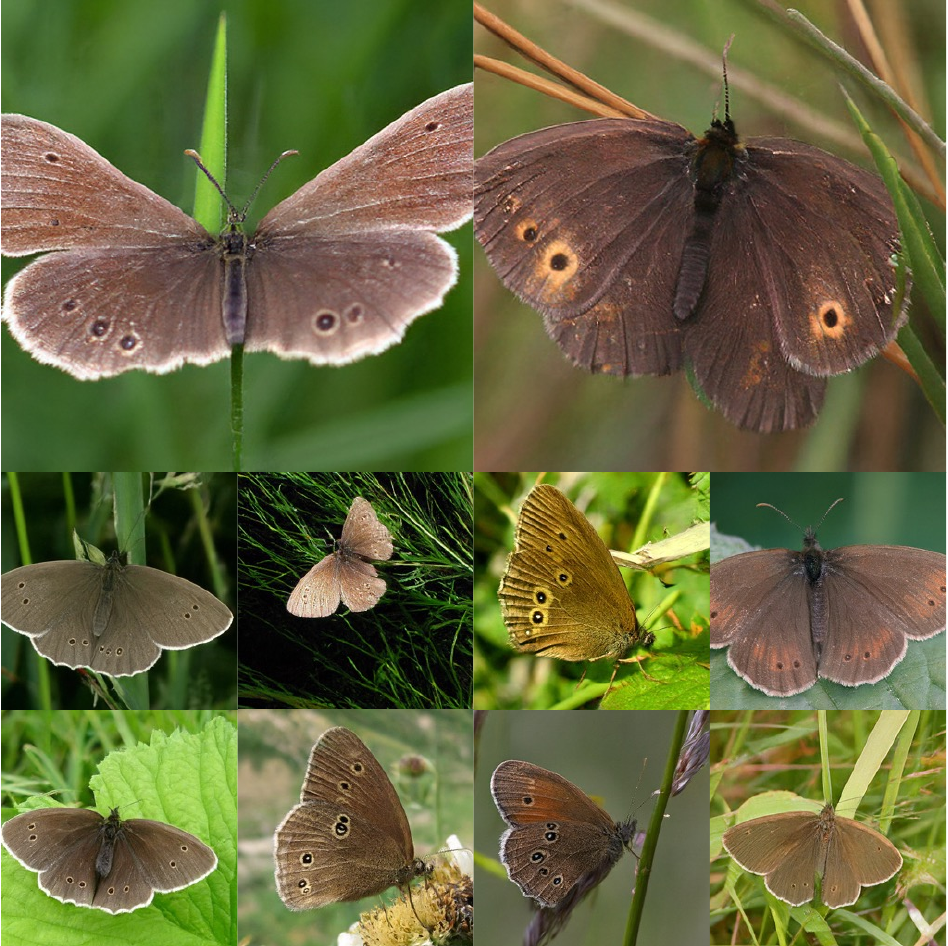}
    \captionof{figure}{ImageNet class 322.}
    \label{fig:app-class-sample-322}
  \end{minipage}\hfill%
  \begin{minipage}[t]{0.48\textwidth}
    \centering
    \includegraphics[width=\linewidth]
      {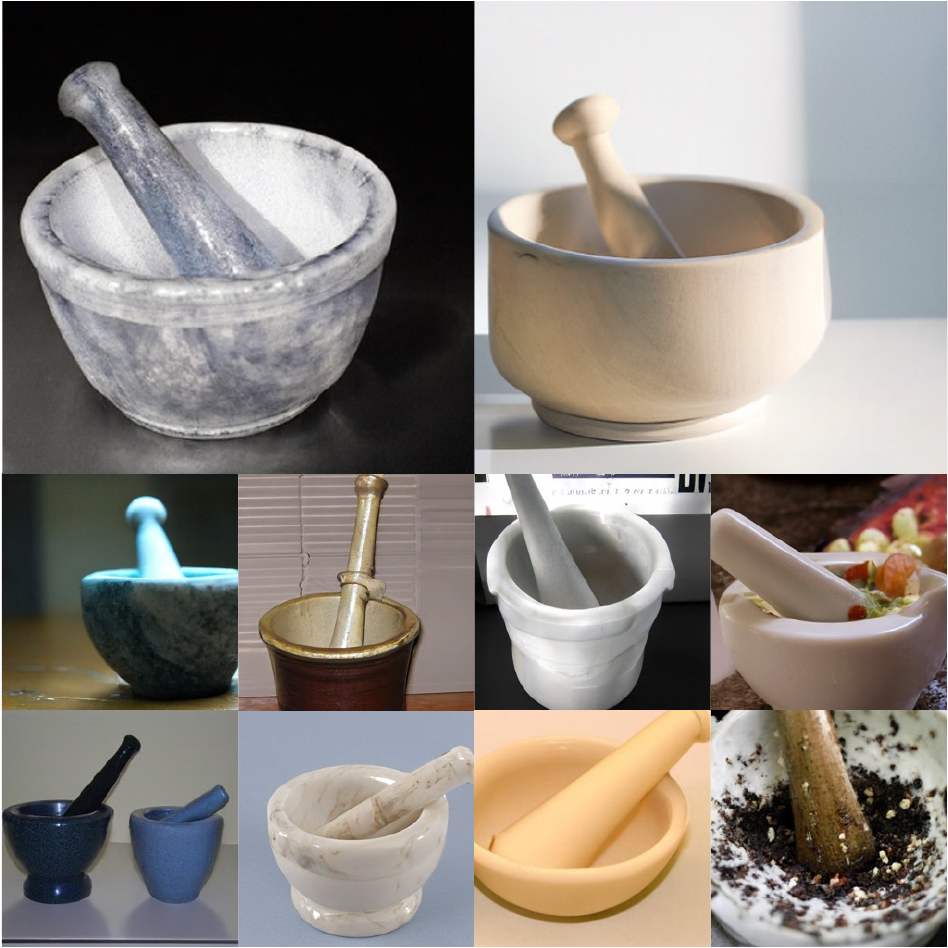}
    \captionof{figure}{ImageNet class 666.}
    \label{fig:app-class-sample-666}
  \end{minipage}
\par\vspace*{\fill}

\clearpage
\vspace*{\fill}
\noindent
  \begin{minipage}[t]{0.48\textwidth}
    \centering
    \includegraphics[width=\linewidth]
      {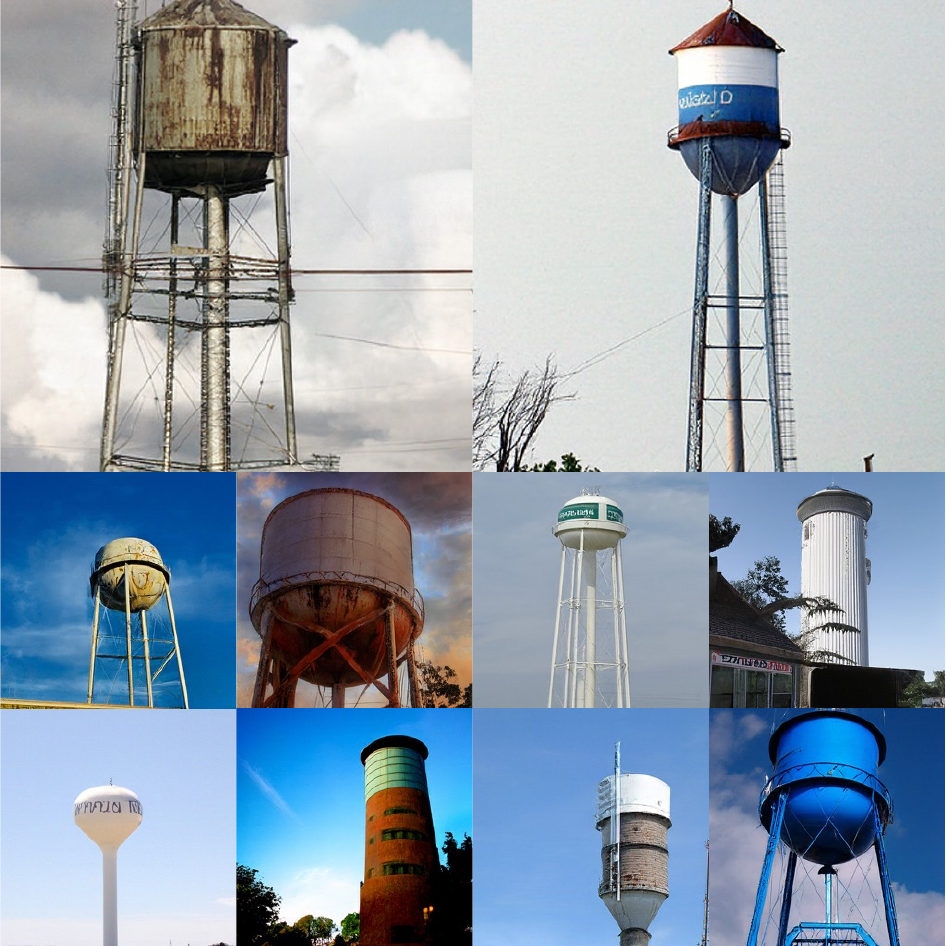}
    \captionof{figure}{ImageNet class 900.}
    \label{fig:app-class-sample-900}
  \end{minipage}\hfill%
  \begin{minipage}[t]{0.48\textwidth}
    \centering
    \includegraphics[width=\linewidth]
      {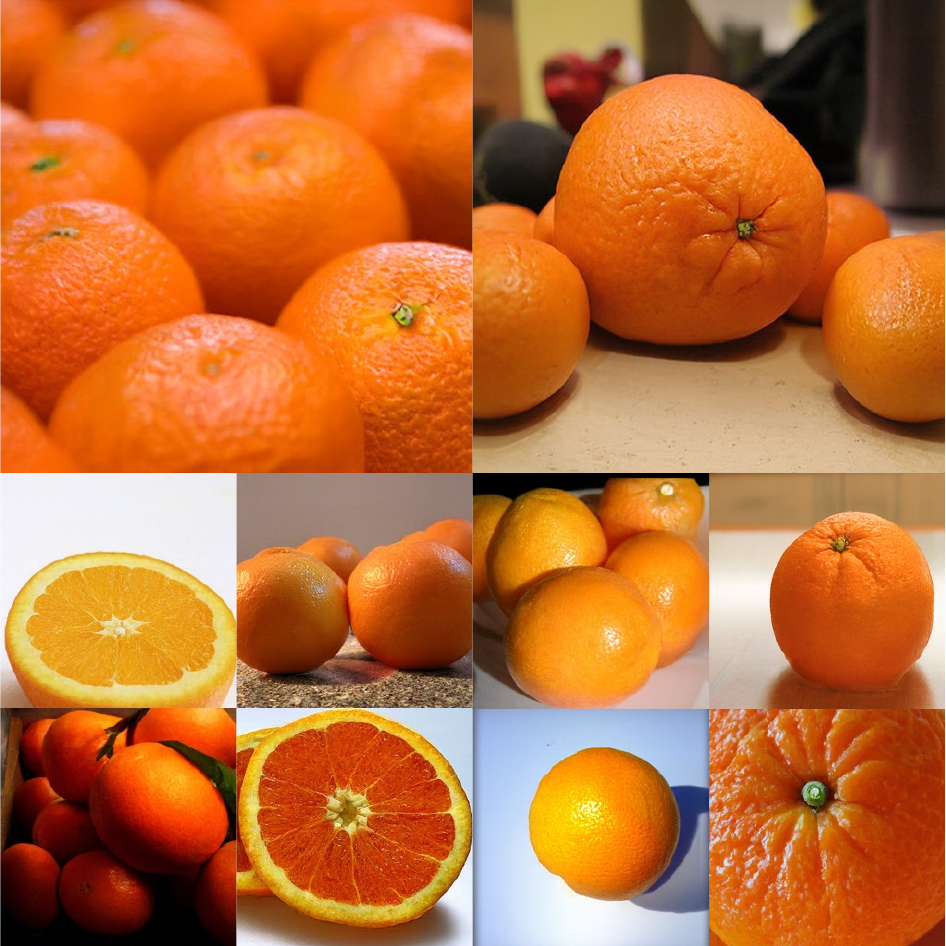}
    \captionof{figure}{ImageNet class 950.}
    \label{fig:app-class-sample-950}
  \end{minipage}

  \par\vspace{0.025\textheight}
\noindent
  \begin{minipage}[t]{0.48\textwidth}
    \centering
    \includegraphics[width=\linewidth]
      {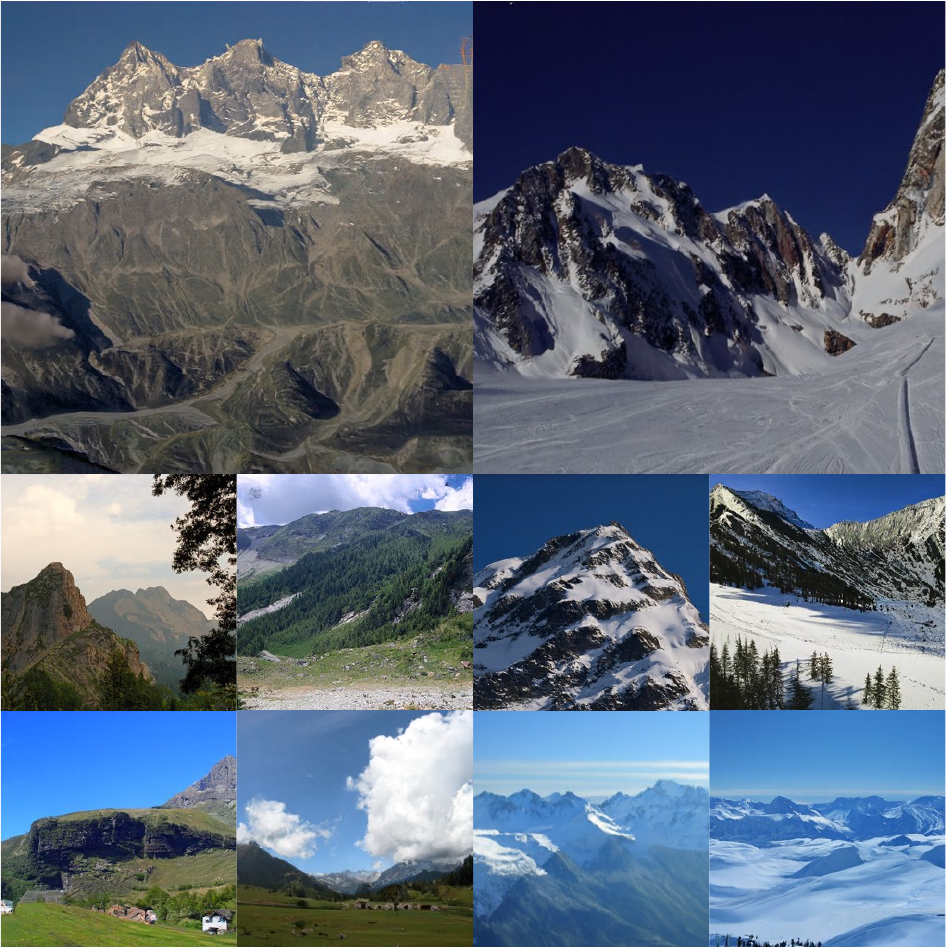}
    \captionof{figure}{ImageNet class 970.}
    \label{fig:app-class-sample-970}
  \end{minipage}
  \hfill%
  \begin{minipage}[t]{0.48\textwidth}
    \centering
    \includegraphics[width=\linewidth]
      {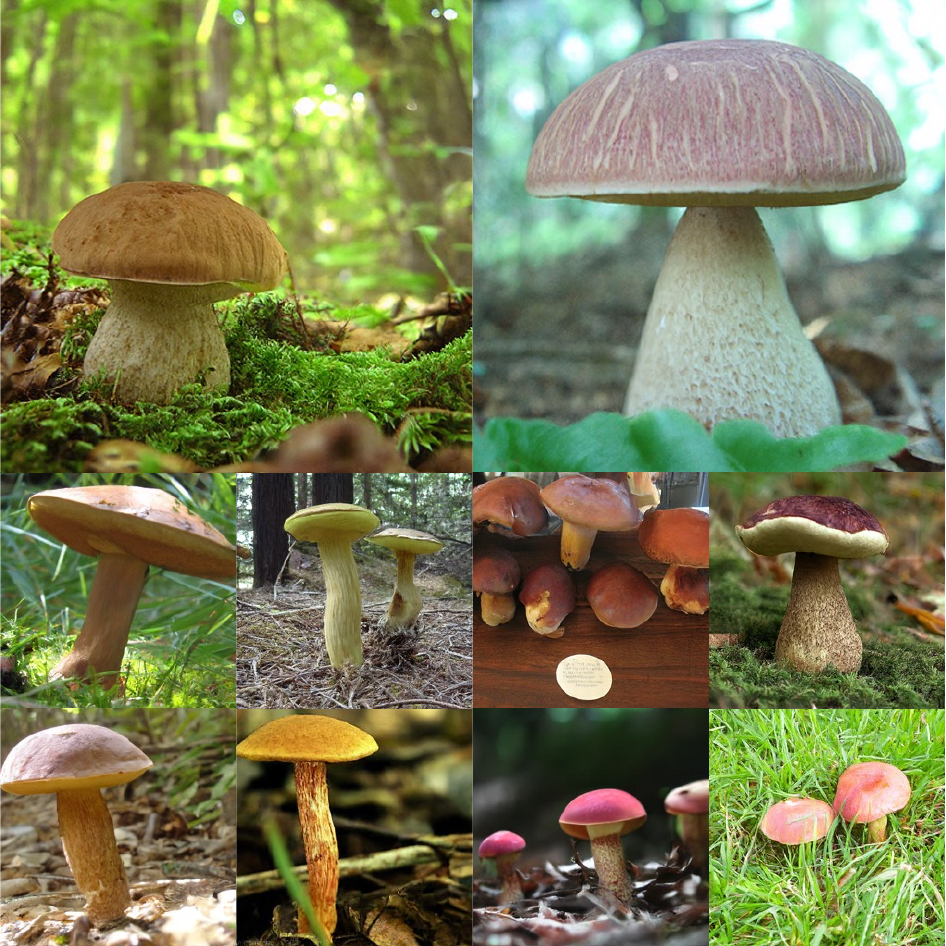}
    \captionof{figure}{ImageNet class 997.}
    \label{fig:app-class-sample-997}
    \end{minipage}
\par\vspace*{\fill}

\end{document}